\documentclass[]{article}

\usepackage[final]{neurips_2022}

\usepackage[utf8]{inputenc} %
\usepackage[T1]{fontenc}    %
\usepackage[hidelinks]{hyperref}       %
\usepackage{url}            %
\usepackage{booktabs}       %
\usepackage{amsfonts}       %
\usepackage{nicefrac}       %
\usepackage{microtype}      %
\usepackage[dvipsnames]{xcolor}         %
\usepackage{natbib}  %
\newlength\mylen
\setcitestyle{numbers,square,citesep={,\kern-\mylen}}
\usepackage{multirow}

\usepackage{tikz}

\usepackage{inconsolata}
\usepackage{caption}
\usepackage{proof}
\usepackage{amsmath,amsfonts,amssymb,verbatim,ifthen,amsthm}
\usepackage{todonotes}  %
\usepackage{dirtytalk}  %
\usepackage{graphicx}
\usepackage{subcaption}  %
\usepackage{lipsum}  %
\usepackage{paralist}  %
\usepackage[multiple]{footmisc}  %
\usepackage{soul}  %
\usepackage{wrapfig}

\usepackage{pifont}%
\newcommand{\cmark}{\ding{51}}%
\newcommand{\xmark}{\ding{55}}%
\newcommand{\com}[1]{}

\usepackage{bbm}
\usetikzlibrary{positioning}

\usepackage{amsthm}
\newtheorem{definition}{Definition}

\newcommand{\cebab}{CEBaB}
\newcommand{\opentable}{OpenTable}

\newcommand{\ATE}{\textit{ATE}}
\newcommand{\doop}{\textit{do}}
\newcommand{\nn}{\mathcal{N}}

\newcommand{\ICACEerror}{\text{ICaCE-Error}}
\usepackage{babel}
\usepackage{rotating}

\definecolor{iem_yellow}{RGB}{244,227,22}
\definecolor{iem_blue}{RGB}{26,37,53}

\title{CEBaB: Estimating the Causal Effects of Real-World Concepts on NLP Model Behavior}

\author{%
  Eldar David Abraham\thanks{Equal contribution. Author names alphabetical.}${}^{\ast,1}$ \\   
  \texttt{eldar.a@campus.technion.ac.il}
  \And
  Karel D'Oosterlink$^{\ast,2, 3}$ \\  
  \texttt{Karel.DOosterlinck@UGent.be}
  \And
  Amir Feder$^{\ast, 1}$ \\  
  \texttt{feder@campus.technion.ac.il}
  \And
  Yair Gat$^{\ast, 1}$ \\
  \texttt{yairgat@campus.technion.ac.il}
  \And
  Atticus Geiger$^{\ast, 2}$ \\
  \texttt{atticusg@stanford.edu} 
  \And
  Christopher Potts$^{\ast, 2}$ \\  
  \texttt{cgpotts@stanford.edu}
  \And
  Roi Reichart$^{\ast, 1}$ \\  
  \texttt{roiri@technion.ac.il} \\
  \And
  Zhengxuan Wu$^{\ast, 2}$ \\  
  \texttt{wuzhengx@stanford.edu} \\
  \AND
  ${}^{1}$Technion -- Israel Institute of Technology 
  \And 
  ${}^{2}$Stanford University
  \And
  ${}^{3}$Ghent University
}

\begin{document}

\maketitle

\begin{abstract}

The increasing size and complexity of modern ML systems has improved their predictive capabilities but made their behavior harder to explain. Many techniques for model explanation have been developed in response, but we lack clear criteria for assessing these techniques. In this paper, we cast model explanation as the causal inference problem of estimating causal effects of real-world concepts on the output behavior of ML models given actual input data. We introduce \cebab, a new benchmark dataset for assessing concept-based explanation methods in Natural Language Processing (NLP). \cebab\ consists of short restaurant reviews with human-generated counterfactual reviews in which an aspect (food, noise, ambiance, service) of the dining experience was modified. Original and counterfactual reviews are annotated with multiply-validated sentiment ratings at the aspect-level and review-level.
The rich structure of \cebab\ allows us to go beyond input features to study the effects of abstract, real-world concepts on model behavior. We use \cebab\ to compare the quality of a range of concept-based explanation methods covering different assumptions and conceptions of the problem, and we seek to establish natural metrics for comparative assessments of these methods.%

\end{abstract}

\section{Introduction}\label{sec:intro}

Explaining model behavior has emerged as a central goal within ML. In NLP, models have grown in size and complexity, and while they have become increasingly successful, they have also become more opaque  \citep{lipton_mythos_2018, pearl_limitations_2019}, raising concerns about trust \citep{guidotti2018survey, jacovi2020towards}, safety \citep{amodei_concrete_2016,otte_safe_2013}, and fairness \citep{goodman_european_2017,hardt_equality_2016}. These concerns will persist if these models remain ``black-boxes''.

Seeking to open the black-box, researchers have developed methods that try to explain model behavior \citep{bastings2021proceedings,feder_causalm_2020,gehrmann_causal_2020,lundberg_unified_2017,ribeiro_why_2016}. 
However, there is no consensus about how to evaluate such methods
to allow
robust comparisons. This is not surprising, since such evaluations require very rich empirical data. Intuitively, we would like to (1) intervene on model inputs, to modify specific concepts without changing other correlated information, (2) observe the effects this has on model predictions, and, finally, (3) 
assess explanation methods for their ability to accurately predict these effects.

The absence of interventional data, or even an agreed-upon non-interventional benchmark, has created an environment in which explanation methods are often evaluated individually, and without comparison to alternatives. Attempts have been made to conduct comparative evaluations \citep{feder_causalm_2020,goyal_explaining_2020, pruthi2022evaluating}, but only with synthetic, simplified datasets. Furthermore, these attempts do not define a unified evaluation approach, nor do they seek to contribute benchmark datasets that support such evaluations.

In this paper, we seek to overcome this obstacle by introducing \textbf{\cebab{}} (\textbf{C}ausal \textbf{E}stimation-\textbf{Ba}sed \textbf{B}enchmark). 
Table~\ref{tab:cebab-toy} summarizes the structure of \cebab\ with a toy example: beginning with a review text from the \opentable\ website, we crowdsourced edits of the original text that are designed to meet a specific goal, such as changing the food rating in the original text to negative or unknown. All of the resulting edits were validated by five crowdworkers and each full text was evaluated by five crowdworkers for its overall sentiment. \cebab\ is grounded in 2,299 original reviews, which were expanded via this editing procedure to a total of 15,089 texts, targeting four different aspect-level concepts (food, service, ambiance, noise) with three potential labels (positive, negative, and unknown, i.e., not expressed in the review), and each full text was labeled on a five-star scale.

We focus on using \cebab\ to compare concept-based explanation methods. This allows us to go beyond the effect of individual tokens to study how more abstract concepts (in our case, aspect-level sentiment) contribute to model predictions (about the overall sentiment of the text). Our proposed metrics center around assessing concept-based explanation methods for their ability to accurately estimate \emph{causal concept effects} \citep{goyal_explaining_2020}, allowing us to isolate the effect of individual concepts.

\begin{table}[tp]
  \caption{Toy examples illustrating the structure of \cebab\ (actual corpus examples are longer and more complex; a sample is given in Appendix~\ref{app:cebab}). Beginning from an \opentable\ review, we give crowdworkers an actual restaurant review and they generate counterfactual restaurant reviews that would have been written if some aspect of the dining experience were changed and all else were held constant. Five different crowdworkers labeled each of the actual and counterfactual texts according to their aspect-level sentiment and overall sentiment. Aspect level sentiment labels are three way: `+' (positive sentiment), `--' (negative), and `unk' (the aspect's value is not expressed in the text). Overall sentiment labels are 1 (worst) to 5 (best). Edited aspect labels are shown in blue.}
  \label{tab:cebab-toy}
  
  \vspace{3pt}
  
  \centering
  \resizebox{\textwidth}{!}{
  \footnotesize
  \setlength{\tabcolsep}{3pt}
  \begin{tabular}[]{@{} r@{ \ }c l c c c c c @{}}
    \toprule
    & & & { food} & {ambiance} & {service} & {noise} & {overall} \\
    \midrule
    \multicolumn{2}{c}{\textbf{Original text:}} &  Excellent lobster and decor, but rude waiter. & + & + & -- & unk & 4  \\
    \midrule
    \multicolumn{2}{c}{\textbf{Edit Goal}}  \\[2ex]
    food: & \textcolor{Cerulean}{--}  & Terrible lobster, excellent decor, but rude waiter. & \textcolor{Cerulean}{--} & + & -- & unk & 2 \\
    food:& \textcolor{Cerulean}{unk} &  Excellent decor, but rude waiter. & \textcolor{Cerulean}{unk} & + & -- & unk & 3
    \\[1ex]
    ambiance:& \textcolor{Cerulean}{--}  &  Excellent lobster, but lousy decor and rude waiter. & + & \textcolor{Cerulean}{--} & -- & unk & 3 \\
    ambiance:& \textcolor{Cerulean}{unk} &  Excellent lobster, but rude waiter. & + & \textcolor{Cerulean}{unk} & -- & unk & 3
    \\[1ex]
    service:& \textcolor{Cerulean}{+} &  Excellent lobster and decor, and friendly waiter. & + & + & \textcolor{Cerulean}{+} & unk & 5 \\
    service:& \textcolor{Cerulean}{unk} & Excellent lobster and decor. & + & + & \textcolor{Cerulean}{unk} & unk & 5
    \\[1ex]
    noise:& \textcolor{Cerulean}{+} &  Excellent lobster, decor, and music, but rude waiter. & + & + & -- & \textcolor{Cerulean}{+} & 4 \\
    noise:& \textcolor{Cerulean}{--} & Excellent lobster and decor, but rude waiter, and noisy. & + & + & -- & \textcolor{Cerulean}{--} & 3 \\
    \bottomrule
  \end{tabular}
  }
  
  \vspace{-20pt}

\end{table}

More specifically, we use \cebab\ to measure the causal effects of particular variables in a causal graph, and we cast each explanation method as a causal estimator of these measurements. For example, suppose our causal graph of the data says that all four of our aspect-level categories will affect a reviewer's overall rating. To estimate the effect of positive food quality on the predicted overall rating from a classifier, we need to compare examples with high food quality to those with low quality, holding all other aspects constant. Such pairs of examples are normally not observed, but this is precisely what \cebab\ provides. With \cebab, we can directly compare the actual change in model predictions with the change that a concept-based explanation method predicts.

In our experiments, we evaluate five leading concept-based explanation methods:
CONEXP \citep{goyal_explaining_2020},
TCAV \citep{kim_interpretability_2018},
ConceptSHAP \citep{yeh2020completeness},
INLP \citep{ravfogel_null_2020},
CausaLM \citep{feder_causalm_2020}, and
{S-Learner} \citep{kunzel2019metalearners}. These methods make a wide range of different assumptions about how much access we have to the model's internal structure, and they also diverge in the degree to which they account for the causal nature of the concept effect estimation problem. Remarkably, \cebab\ reveals that most methods cannot beat a simple baseline. Indeed, this negative result emphasizes the value in our primary contribution of providing the data and metrics that enables a direct comparison of explanation methods.

\section{Previous Work}\label{sec:prev}

\paragraph{Benchmarks for Explanation Methods}\label{subsec:benchmarks}
Benchmark datasets have propelled ML forward by creating shared metrics that predictive models can be evaluated on \citep{hu2020xtreme,kiela2021dynabench,wang_glue_2018,wang2019superglue}. Unfortunately, benchmarks that are suitable for assessing the quality of model explanations are still uncommon \citep{feder_causal_2021,hooker2019benchmark}. Previous work on comparing explanation methods has generally only correlated the performance of a given explainability method with others, without ground-truth comparisons \citep{deyoung2020eraser, hase2020evaluating,hooker2019benchmark,samek2021explaining}.

Other works that do compare to some ground-truth either employ a non-causal evaluation scheme \citep{kim_interpretability_2018}, use causal evaluation metrics which do not capture performance on individual examples~\citep{tenney-etal-2020-language}, evaluate on synthetic counterfactuals and rule-based augmentations \citep{feder_causalm_2020, tenney-etal-2020-language}, or are tailored for a specific explanation method and hard to generalize \citep{yeh2020completeness}. To the best of our knowledge, CEBaB is the first large-scale naturalistic causal benchmark with interventional data for NLP.

\paragraph{Explanation Methods and Causality}\label{subsec:methods}Probing is a relatively new technique for understanding what model internal representations encode. In probing, a small supervised  \citep{conneau-etal-2018-cram,tenney-etal-2019-bert} or unsupervised  \citep{clark-etal-2019-bert,Manning-etal:2020,saphra-lopez-2019-understanding} model is used to estimate whether specific concepts are encoded at specific places in a network. While probes have helped illuminate what models (especially pretrained ones) have learned from data, \citet{geiger2021causal} show with simple analytic examples that probes cannot reliably provide causal explanations for model behavior.

Feature importance methods can also be seen as explanation methods \citep{molnar2020interpretable}. Many methods in this space are restricted to input features, but gradient-based methods can often quantify the relative importance of hidden states as well \citep{Binder16,Shrikumar16,springerberg2014,Zeiler2014}. The Integrated Gradients method of \citet{sundararajan17a} has a natural causal interpretation stemming from its exploration of baseline (counterfactual) inputs \citep{geiger2021causal}. However, even where these methods can focus on internal states, it remains difficult to connect their analyses with real-world concepts that do not reduce to simple properties of inputs.

Intervention-based methods involve modifying inputs or internal representations and studying the effects that this has on model behavior \citep{lundberg_unified_2017,ribeiro_why_2016}. Recent methods perturb input or hidden representations to create counterfactual states that can then be used to estimate causal effects \citep{elazar2021amnesic,finlayson-etal-2021-causal,soulos-etal-2020-discovering,vig2020causal, geiger2021causal}. However, these methods are prone to generating implausible inputs or network states unless the interventions are carefully controlled \citep{geiger-etal-2020-neural}.

Generating counterfactual texts automatically remains challenging and is still a work-in-progress \citep{calderon2022docogen}. 
To overcome this problem, another class of approaches proposes to manipulate the representation of the text with respect to some concept, rather than the text itself \citep{elazar2021amnesic,feder_causalm_2020,ravfogel_null_2020}. These methods fall into the category of concept-based explanations and we discuss two of them extensively in \S\ref{sec:methods}.

\section{Estimating Concept Effects with \cebab}\label{sec:methods}

\begin{figure}[tp]
\centering    
        \begin{tikzpicture}[node distance = 2.0cm]
            \node[text centered] (O) {$\nn (\phi(X))$};
            \node[left of =  O] (I) {$X$};
            \node[left of = I] (C2) {$C_2$};
            \node[above = 0.25cm of C2] (C1) {$C_1$};
            \node[below =-0.1cm of C2] (Dot) {$\Large \vdots$};
            \node[below =0.75cm of C2] (Ck) {$C_k$};
            \node[left of = C2] (U) {$U$};
            \node[above = 0.75cm of I] (V) {$V$};
    
            \draw[->, line width = 1] (I) -- (O);
            \draw[->, line width = 1] (C1) -- (I);
            \draw[->, line width = 1] (C2) -- (I);
            \draw[->, line width = 1] (Ck) -- (I);
            \draw[->, line width = 1, dashed] (U) -- (C1);
            \draw[->, line width = 1,dashed] (U) -- (C2);
            \draw[->, line width = 1,dashed] (V) -- (I);
            \draw[->, line width = 1,dashed] (U) -- (Ck);
        \end{tikzpicture}
    \caption{A causal graph describing a data generating process with an exogenous variables $U$ and $V$ representing the state of the world, mediating concepts $C_1, C_2 \dots, C_k$, and data $X$ that is featurized with $\phi$. $\phi$($X$) is input to a classifier $\nn$, which outputs a vector of scores over $m$ output classes.}
    \label{fig:general-graph}
\end{figure}
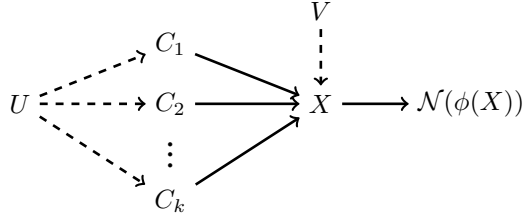

We now define the core metrics that we use to evaluate different explanation methods. Figure~\ref{fig:general-graph} provides a high-level view of the causal process we are envisioning. The process begins with an exogenous variable $U$ representing a state of the world. For \cebab, we can imagine that the value of $U$ is a state of affairs $u$ of a person evaluating a restaurant in a particular way. $u$ contributes to a review variable $X$, with the value $x$ of $X$ mediated by $u$ and by mediating concepts $C_{1}, \ldots C_{k}$, which correspond to the four aspect-level categories in \cebab\ (food, service, ambiance, and noise), each of which can have values $c \in \{\text{positive}, \text{negative}, \text{unknown}\}$. The review $x$ is processed by a model that outputs a vector of scores over classes (sentiment labels in \cebab).

\paragraph{Core Metric} Our central goal is to use \cebab\ to evaluate explanation methods themselves. \cebab\ supports many approaches to such evaluation. In this paper, we adopt an approach based on individual-level rather than average effects. This makes very rich use of the counterfactual text and associated labels provided by \cebab. 
The starting point for this metric is the Individual Causal Concept Effect:
\begin{definition}[Individual Causal Concept Effect; ICaCE]\label{def:icace}
For a neural network $\nn$ and feature function $\phi$, the individual causal concept effect of changing the value of concept $C$ from $c$ to $c'$ for state of affairs $u$ in an underlying data generation process $\mathcal{G}$ is
    \begin{equation}
        \text{ICaCE}_{\mathcal{N}_{\phi}}(\mathcal{G}, x_u^{C=c}, c') =
        \mathbb{E}_{x \sim \mathcal{G}}\left[ \nn \big(\phi(x)\big) 
        \, \big| \, 
        \doop\negthinspace
        \left(       
        \begin{array}{@{} c @{}}
        C=c' \\
        U=u
        \end{array}
        \right)
    \right]
            -
         \nn \big(\phi(x^{C=c}_u)\big).
    \end{equation}
\end{definition}
ICaCE is a theoretical quantity. In practice, we use the Empirical Individual Causal Concept Effect.
\begin{definition}[Empirical Individual Causal Concept Effect; $\widehat{\text{ICaCE}}$]\label{def:eicace}
For a neural network $\nn$ and feature function $\phi$, the empirical individual causal concept effect of changing the value of concept $C$ from $c$ to $c'$ for state of affairs $u$ is
\begin{equation}   
    \label{eq:eicace}
    \widehat{\text{ICaCE}}_{\nn_{\phi}}(x^{C=c}_u, x^{C=c'}_u) =  
    \nn \big(\phi(x^{C=c'}_u)\big)
    -
    \nn \big(\phi(x^{C=c}_u)\big),
\end{equation}
where $(x^{C=c}_u, x^{C=c'}_u)$ is a tuple of inputs originating from $u$ with the concept $C$ set to the values $c$ and $c'$, respectively.
\end{definition}
The $\widehat{\text{ICaCE}}_{\nn_{\phi}}$ for a pair of examples $(x^{C=c}_u, x^{C=c'}_u)$ is simply the difference between the output score vectors for the two cases. With \cebab, we can easily calculate these values because we have clusters of examples that are tied to the same reviewing situation $u$ and express different concept values.

For assessing an explanation method $\mathcal{E}$, we compare ICaCE values with those returned by $\mathcal{E}$. Our core metric is the ICaCE-Error:
\begin{definition}[\ICACEerror]\label{def:ICaCE-Error}
For a neural network $\nn$, feature function $\phi$ and distance metric $\mathsf{Dist}$, the ICaCE-Error of an explanation method $\mathcal{E}$ for changing the value of concept $C$ from $c$ to $c'$ is:
\begin{equation}\label{eq:ICaCE-Error}
    \ICACEerror_{\nn_{\phi}}^{\mathcal{D}}(\mathcal{E}) = 
    \frac{1}{\left|\mathcal{D}\right|} 
    \negthickspace
    \sum_{(x^{C=c}_u, x^{C=c'}_u) \in \mathcal{D}}
    \negthickspace 
    \! \! \! \!\mathsf{Dist}\big(
    \widehat{\text{ICaCE}}_{\nn_{\phi}}(x^{C=c}_u, x^{C=c'}_u), 
    \mathcal{E}_{\nn_{\phi}}(x^{C=c}_u, c')
    \big)
\end{equation}
\end{definition}
We present results for three choices of $\mathsf{Dist}$ which vary in their ability to model the direction and magnitude of effects. 
These choices give subtly different but largely converging results, as detailed in Section~\ref{sec:exp} and reported more fully in Appendix~\ref{app:extended-results}.

\paragraph{Aggregating Individual Causal Concept Effect}
It is often useful to also have a direct estimate of a model's ability to capture concept-level causal effects. For this, we employ an aggregating version of $\widehat{\text{ICaCE}}$, the Empirical Causal Concept Effect:
\begin{definition}[Empirical Causal Concept Effect; $\widehat{\text{CaCE}}$]\label{def:empirical-cace}
For a neural network $\nn$ and feature function $\phi$, the empirical causal concept effect of changing the value of concept $C$ from $c$ to $c'$ in dataset $\mathcal{D}$ is
\begin{equation}\label{def:CaCE}
    \widehat{\text{CaCE}}_{\nn_{\phi}}^{\mathcal{D}}(C, c, c') = \frac{1}{|\mathcal{D}_{C}^{c \to c'}|} \sum_{(x^{C=c}_u, x^{C=c'}_u) \in \mathcal{D}_{C}^{c \to c'}} \widehat{\text{ICaCE}}_{\nn_{\phi}}(x^{C=c}_u, x^{C=c'}_u).
\end{equation}
\end{definition}
This is an empirical estimator of the Causal Concept Effect (CaCE) of \citet{goyal_explaining_2020}. It estimates, in general, how the classifier predictions change for a given concept and intervention direction.

\paragraph{Estimating Real-World Causal Effect of Aspect Sentiment on Overall Sentiment}\label{sec:ite}
We can also estimate ground truth causal effects in \cebab\ by simply using its labels directly. There are again a variety of ways that this could be done. We opt for the one that makes the richest use of the structures afforded by \cebab. For perspicuity, in parallel to the neural network-based $\widehat{\text{ICaCE}}$ (Definition~\ref{def:eicace}), we define the Empirical Individual Treatment Effect for our dataset:
\begin{definition}[Empirical Individual Treatment Effects in \cebab; $\widehat{\text{ITE}}$]\label{def:ite}
The empirical individual treatment effect of changing the value of concept $C$ from $c$ to $c'$ in \cebab\ is
\begin{equation}\label{eq:ATE}
    \widehat{\text{ITE}}^{\text{\cebab}}(x^{C=c}_u, x^{C=c'}_u) =     
    f(x^{C=c'}_u) - f(x^{C=c}_u)    
    \end{equation}
    where $f$ is a simple look-up procedure that retrieves the overall sentiment labels for \cebab\ examples.
\end{definition}
We aggregate over these values by taking their average, in parallel to what we do for network predictions (Definition~\ref{def:empirical-cace}). This yields the Empirical Average Treatment Effect ($\widehat{\text{ATE}}$) for \cebab.

\paragraph{Alternative Metrics}
In Appendix~\ref{app:theory} in our supplementary materials, we consider alternative formulations of the core metrics with \textit{causal concept effects} and \textit{absolute causal concept effects}, relating them to the different questions they engage with. We opt for the individual causal concept effect in our central metric (Definition~\ref{eq:ICaCE-Error}), taking the central question to be what caused an ML model to produce an output for an \textit{actual} input created from a real-world process.

\section{Evaluated Explanation Methods}\label{subsec:explanation-methods}
We compare several model explanation methods that share three main characteristics. First, they are all suitable for NLP models and have been used in the literature for generating model explanations in the form of estimated effects on model predictions. Second, they all provide concept-level explanations, for a pre-defined list of human-interpretable concepts (e.g., how sensitive a restaurant review rating classifier is to language related to food quality). This approach is also forward-looking, allowing more researchers to construct new hypotheses (i.e., concepts we have not collected labels for) and estimate their effect on the predictor.
Third, all of the tested methods are model-agnostic, meaning that they separate the explanation from the model. 
At the same time, these methods differ in five important ways, as summarized Table~\ref{tab:methods-summary}. 

\bgroup
\begin{table}[tp]
    \caption{The evaluated explanation methods and their attributes. \textbf{Explainer Method} denotes the complexity of the models used by each explanation method. \textbf{Access to Explained Model} denotes the degree of access an explainer method needs to the explained model. \textbf{Concept Labels Needed} indicates whether a method estimating the effect for an input $x^{C=c}_u$ needs the actual input label $c$ and/or the intervened value $c'$ at test time. Models with a \textbf{Counterfactual Representation} approximate $\phi(x^{C=c'}_u)$ to estimate the effect. Finally, only CausaLM and S-Learner have \textbf{Confounder Control} to minimize the impact of confounding concepts. ${}^{\ast}$We predict these labels with a classifier.}    
    \label{tab:methods-summary}
    
    \vspace{3pt}
    
    \centering
    \small
    \setlength{\tabcolsep}{5pt}
    \resizebox{\columnwidth}{!}{
        \begin{tabular}{l @{\hspace{12pt}} ccccc}
        \toprule    
        & \textbf{Explainer} 
        & \textbf{Access to }
        & \textbf{Concept Labels}
         & \textbf{Counterfactual} 
         & \textbf{Confounder} \\
        \textbf{Explanation method} 
        & \textbf{Method}
         & \textbf{Explained Model } 
        & \textbf{Needed (test time)} 
        & \textbf{Representation} 
        & \textbf{Control} \\
        \midrule
        Approx & None & None & All concepts and their labels${^\ast} $ & \xmark & \xmark \\
        CONEXP \citep{goyal_explaining_2020}   & None & None & $c$ and $c'$ & \xmark & \xmark \\
        S-Learner \cite{kunzel2019metalearners} & Linear & None & All concepts and their labels${^\ast}$ &\xmark & \cmark \\
        TCAV \citep{kim_interpretability_2018}  & Linear & Weights & None &\xmark & \xmark \\
        ConceptSHAP \citep{yeh2020completeness} & Linear & Weights & None &\xmark & \xmark \\        
        INLP \citep{ravfogel_null_2020}         & Linear & Weights & None &\cmark & \xmark \\
        CausaLM \citep{feder_causalm_2020}      & Explained Model & Training Regime& None &\cmark & \cmark \\
        \bottomrule
    \end{tabular}
    }    
\end{table}
\egroup

We now turn to reviewing the explanation methods that we later compare on \cebab{} (\S \ref{sec:exp}). In our mathematical formulas, we employ a unified notation for all methods, to make the definitions more accessible and easier to integrate into our experimental set-up.
Assume we have a classifier $\nn$ (which outputs a probability vector) and feature function $\phi$, and we want to compute the effect on $\nn_{\phi}(x^{C=c}_u)$ of changing the value of concept $C$ from $c$ to $c'$ using an unseen test set $(\mathcal{D}, Y)$.

\paragraph{Approximate Counterfactuals}
The gold labels of \cebab\ are the difference between the logits for some original review ${x}_{u}^{C=c}$ and ground-truth counterfactual ${x}_{u}^{C=c'}$. As a baseline, we sample an original review ${x}_{u'}^{C=c'}$ with the same aspect-labels as the ${x}_{u}^{C=c'}$ and use it as an approximate counterfactual:
\begin{equation}\label{eq:approx}
    \text{Approx}_{\nn_{\phi}}(C, c, c'; x) = 
    \nn (\phi({x}_{u'}^{C=c'})) -
    \nn (\phi({x}_{u}^{C=c}))
\end{equation}
We do this sampling using predicted aspect labels from the aspect-level sentiment analysis models described in Appendix~\ref{app:cebab-models}.

\paragraph{Conditional Expectation (CONEXP)} 

\citet{goyal_explaining_2020} propose a baseline where the effect of a concept $C$ is the average difference in predictions on examples with different values of $C$.
\begin{equation}\label{eq:conexp}
    \text{CONEXP}_{\nn_{\phi}}^{\mathcal{D}}(C, c, c') = \frac{1}{|\mathcal{D}^{C=c'}|} 
    \sum_{x \in \mathcal{D}^{C=c'}} \nn (\phi(x)) -
    \frac{1}{|\mathcal{D}^{C=c}|} 
    \sum_{x \in \mathcal{D}^{C=c}} \nn (\phi(x))
\end{equation}
where $\mathcal{D}^{C=c}$ and $\mathcal{D}^{C=c'}$ are subsets of $\mathcal{D}$ where $C$ takes values $c$ and $c'$, respectively. To predict an effect, this method only relies on $C$, $c$, and $c'$, resulting in an estimate that does not depend on the specific input text itself.

\paragraph{Conditional Expectation Learner (S-Learner)}
We adapt \textit{S-Learner}, a popular method for estimating the Conditional Average Treatment Effect (CATE) \citep{kunzel2019metalearners}. 
To estimate causal concept effects, our \textit{S-Learner} trains a logistic regression model $\mathcal{E}$ to predict $\nn (\phi(x))$ using the values of all the labeled concepts of example $x$, denoted by $x'$.\footnote{This training approach, where an explainer model is fit to predict the output of the original model, shares the intuition of LIME, the widely used explanation method \cite{ribeiro_why_2016}, but for concept-level effects.} Then, during inference, we compute an individual effect for example pair $({x}_{u}^{C=c}, {x}_{u}^{C=c'})$ by comparing the output of the model $\mathcal{E}_x$ on this pair:
\begin{equation}
    \label{eq:s-learner}
    \textit{S-Learner}(C,c,c';x) = \mathcal{E}({x'}_{u}^{C=c'}) - \mathcal{E}({x'}_{u}^{C=c})
\end{equation}

At inference time, S-Learner assumes access to all aspect-level labels $x'$, which might not always be available. To alleviate this issue, we instead \emph{predict} the aspect-level labels $x'$ from the original text $x$ using models described in Appendix~\ref{app:cebab-models}.

\paragraph{TCAV} 

\citet{kim_interpretability_2018} use \textit{Concept Activation Vectors} (CAVs), which are semantically meaningful directions in the embedding space of $\phi$. Our adapted version of Testing with CAVs (TCAV) outputs a vector measuring the sensitivity of each output class $k$ to changes towards the direction of a concept $v_C$ at the point of the embedded input. It is computed as:
\begin{equation}\label{eq:tcav_score}
    \text{TCAV}_{\nn_{\phi}}(C; x)
    =
    \big(\nabla \nn_k (\phi(x)) \cdot v_C\big)_{k=1}^K
\end{equation}
where $K$ is the number of classes and $v_C$ is a linear separator learned to separate concept $C$ in the embedding space of $\phi$.

\paragraph{ConceptSHAP} 

\citet{yeh2020completeness} propose this expansion to SHAP \citep{lundberg_unified_2017}, to generate concept-based explanation based on Shapley values \citep{shapley1953quota}.
Given a \textit{complete} (i.e., such that the accuracy it achieves on a test set is higher than some threshold $\beta$) set of $m$ concepts $\{ C_1, \ldots, C_m\}$, ConceptSHAP calculates the contribution of each concept to the final prediction. Our adapted version outputs a vector for each $C \in \{ C_1, \ldots, C_m\}$ and $x$. We justify this modification and provide implementation details in Appendix~\ref{app:concept-shap}.

\paragraph{CausaLM} 

\citet{feder_causalm_2020} estimate the causal effect of a binary concept $C$ on the model's predictions by adding auxiliary adversarial tasks to the language representation model in order to learn a counterfactual representation $\phi^{\text{CF}}_C(x)$, while keeping essential information about potential confounders (control concepts). Their method outputs the text representation-based individual treatment effect (TReITE), which is computed as:
\begin{equation}\label{eq:TReATE_hat_O_CF}
    \text{TReITE}_{\nn_{\phi}}(C; x) = 
    \nn' \big(\phi^{\text{CF}}_C(x)\big) -  \nn \big(\phi(x)\big) 
\end{equation}
where $\phi^{\text{CF}}_C$ denotes the learned counterfactual representation, where the information about concept $C$ is not present, and $\nn'$ is a classifier trained on this counterfactual representation. A key feature of CausaLM is its ability to control for confounding concepts (if modeled).\footnote{As in \citet{feder_causalm_2020}, we control for the most correlated potential confounder.}
An inherent drawback of this technique is that it can only estimate interventions well for $c' = \text{Unknown}$, since the counterfactual representation is only trained to \emph{remove} a concept $C$.

\paragraph{Iterative Nullspace Projection (INLP)} 

\citet{ravfogel_null_2020} remove a concept from a representation vector by repeatedly training linear classifiers that aim to predict that attribute from the representations and projecting the learned representations on their null-space. Similar to CausaLM, INLP also estimates the TReATE (Equation~\ref{eq:TReATE_hat_O_CF}) and can only estimate interventions for $c' =  \text{Unknown}$.

\section{The \cebab\ Dataset}\label{sec:data}

\begin{table}[tp]
  \caption{Dataset statistics of \cebab\ combining train/dev/test splits.}
  \label{tab:cebab-label-dists}
    
  \vspace{3pt}

  \centering
  \begin{subtable}[b]{0.7\linewidth}
    \centering
    \setlength{\tabcolsep}{3pt}
    \begin{tabular}{@{} l *{4}{r@{ \ }r} r @{}}
      \toprule
      {} &  \multicolumn{2}{c}{Positive} &  \multicolumn{2}{c}{Negative} & \multicolumn{2}{c}{Unknown} & \multicolumn{2}{c}{no maj.} & Total \\
      \midrule
      food     &      5726 &       (41\%) &      5526 &       (38\%) &     2605 &      (15\%) &          208 &          (31\%) &  14065 \\
      service  &      4045 &       (29\%) &      4098 &       (28\%) &     3877 &      (22\%) &          178 &          (27\%) &  12198 \\
      ambiance &      2928 &       (21\%) &      2597 &       (18\%) &     5121 &      (29\%) &          203 &          (30\%) &  10849 \\
      noise    &      1365 &       (10\%) &      2215 &       (15\%) &     5883 &      (34\%) &           78 &          (12\%) &   9541 \\
      \bottomrule
    \end{tabular}
    \caption{Aspect-level labels.}
    \label{tab:cebab-label-dists-aspect}
  \end{subtable}  
  \hfill
  \begin{subtable}[b]{0.28\linewidth}
    \centering
    \setlength{\tabcolsep}{2pt}
    \begin{tabular}{@{} l r@{ \ }r @{}}
      \toprule
      1 star          &            1870 &           (12\%) \\
      2 star           &            3056 &           (20\%) \\
      3 star           &            3517 &           (23\%) \\
      4 star           &            2035 &           (13\%) \\
      5 star           &            2732 &           (18\%) \\
      no maj. &            1879 &           (12\%) \\
      \bottomrule
    \end{tabular}
    \caption{Review-level ratings.}
    \label{tab:cebab-label-dists-review}
  \end{subtable}
  
  \vspace{6pt}
  \begin{subtable}[t]{0.48\linewidth}
    \centering
    \setlength{\tabcolsep}{2pt}
    \begin{tabular}{@{} lrrr @{}}
    \toprule
    {} &  \{Neg, Pos\} &  \{Neg, Unk\} &  \{Pos, Unk\} \\
    \midrule
    food     &                   898 &                 1316 &                 1291 \\
    service  &                   851 &                  857 &                  938 \\
    ambiance &                   947 &                  585 &                  472 \\
    noise    &                  1145 &                  208 &                  260 \\
    \bottomrule
    \end{tabular}   
    \caption{Edit pair distribution. Edit pairs are examples that come from the same original source text and differ only in their rating for a particular aspect.}
    \label{tab:cebab-edit-pairs}
  \end{subtable}
  \hfill
  \begin{subtable}[t]{0.48\linewidth}
  \centering
  \setlength{\tabcolsep}{2pt}
  \begin{tabular}{l c c c}
    \toprule
     & Neg to Pos & Neg to Unk & Pos to Unk \\
    \midrule
    food      &                 1.84 &                1.37 &               $-$1.02 \\
    service   &                 0.98 &                0.91 &               $-$0.53 \\
    ambiance  &                 0.93 &                0.91 &               $-$0.50 \\
    noise     &                 0.72 &                0.48 &               $-$0.47 \\
    \bottomrule
    \end{tabular}
  \caption{Empirical $\widehat{\text{ATE}}$ for the five-way sentiment labels in \cebab.  The reverse of a given concept change is the negative of the value given -- e.g., the $\widehat{\text{ATE}}$ for `Pos to Neg' for food is $-$1.84. See Appendix~\ref{app:cebab} for the corresponding values for binary sentiment.}
  \label{tab:cebab-ate}
  \end{subtable}
\end{table}

Table~\ref{tab:cebab-toy} provides an intuitive overview of the structure of \cebab. In the \textit{editing} phase of dataset creation, crowdworkers modified an existing \opentable{}\ review in an effort to achieve a specific aspect-level goal while holding all other properties of the original text constant. Our aspect-level categories are food, ambiance, service, and noise. In the \textit{validation} phrase, crowdworkers labeled each example relative to each aspect as `Positive', `Negative', or `Can't tell' (Unknown). Having five labels per example allows us to infer a majority label or reason in terms of the full label distributions. In the \textit{rating} phase, each full text was labeled using a common five-star scale, again by five crowdworkers.

We began with 2,299 original reviews from \opentable{}\ (related to 1,084 restaurants) and expanded them, via the above editing procedure, into a total of 15,089 texts. The distribution of normalized edit distances has peaks around 0.28 and 0.77, showing that workers made non-trivial changes to the originals, and even often had to make substantial changes to achieve the editing goal. (See Appendix~\ref{app:cebab} for the full distribution.) 

Table~\ref{tab:cebab-label-dists} summarizes the resulting label distributions, where an example has label $y$ if at least 3 of the 5 labelers chose $y$, otherwise it is in the `no majority' category. 99\% of aspect-level edits have a majority label that corresponds to the editing goal, and 88\% of the texts have a review-level majority label on the five-star scale. Overall, these percentages show that workers were extremely successful in achieving their editing goals and that edits have systematic effects on overall sentiment.

The central goal of \cebab{}\ is to create \emph{edit pairs}: pairs of examples that come from the same original text and differ only in their labels for a particular aspect. For example, in Table~\ref{tab:cebab-toy}, the first two `food edit' cases form an edit pair, since they come from the same original text and differ only in their food label. Original texts can also contribute to edit pairs; the original text in Table~\ref{tab:cebab-toy} forms an edit pair with each of the texts it is related to by edits. Table~\ref{tab:cebab-edit-pairs} summarizes the distribution of edit pairs, and Table~\ref{tab:cebab-ate} reports the ground-truth $\widehat{\text{ATE}}$ values (\S\ref{sec:ite}).

We release the dataset with fixed train/dev/test splits. In creating these splits, we enforce two high-level constraints. The first is our `grouped' requirement: for each original review $t$, all texts that are related to $t$ via editing occur in the same split as $t$. This ensures that models are not evaluated on examples that are related by editing to those they have seen in training. Second, if any text $t$ in a group 
received a `no majority' label, then the entire group containing $t$ is put in the train set. This ensures that there is no ambiguity about how to evaluate models on dev and test examples. 

Once these high-level conditions were imposed, the examples were sampled randomly to create the splits. This allows that individual workers can contribute edited texts across splits. This minor compromise was necessary to ensure that we could have large dev and test splits. Appendix~\ref{app:cebab-models} in our supplementary materials shows that worker identity has negligible predictive power.

There are two versions of the train set: \emph{inclusive} and \emph{exclusive}. The inclusive train set  contains all original and edited non-dev/test texts (11,728 texts). The exclusive version samples exactly one train text from each set of texts that are related by editing (1,755 examples). The rationale is that models trained with an original review as well as its edited counterparts may explicitly learn causal effects trivially by aggregating learning signals across inputs. Our exclusive train split prevents this, which helps facilitate fair comparisons between explanation methods and better resembles a real-world setting.

Our dataset is released publicly in JSON format and is available in the Hugging Face \texttt{datasets} library. It includes  restaurant metadata, full rating distributions, and anonymized worker ids. Appendix~\ref{app:cebab} in our supplementary materials provides additional details on the dataset construction, including the prompts used by the crowdworkers, the number of workers per task, worker compensation, and a sample of examples with ratings to help convey the nature of workers' edits and the overall quality of the resulting texts and labels. In addition, Appendix~\ref{app:cebab-models} reports on a wide range of classifier experiments at the aspect-level and text-level that show that models perform well on \cebab\ classification tasks, which bolsters the claim that \cebab\ is a reliable tool for assessing explanation methods.

\section{Experiments and Results}\label{sec:exp}

For each experiment, we fine-tune a pretrained language model to predict the overall sentiment of all restaurant reviews from our \textit{exclusive} \opentable{} train set. Since the goal of our work is not to achieve state-of-the-art performance, but rather to compare explanation methods and demonstrate the usage of \cebab{}, we test the ability of methods to explain commonly used models, trained with standard experimental configurations. 

In the main text, we report results for \texttt{bert-base-uncased} fine-tuned as a five-way classifier. Appendix \ref{app:extended-results} includes results for \texttt{GPT-2}, \texttt{RoBERTa}, and an \texttt{LSTM}, fine-tuned on binary, 3-way and 5-way versions of the sentiment task. All results, including the ground-truth effect that depends on the specific instance of a model, are averaged across 5 seeds.

\bgroup
\begin{table}[tp]
    \caption{$\widehat{\text{CaCE}}$ (Definition~\ref{def:CaCE}) for \texttt{bert-base-uncased} fine-tuned as a 5-way sentiment classifier. Rows are concepts, columns are real-world concept interventions, and each entry indicates the average change in classifier output when the concept is intervened on with the given direction.\protect\footnotemark\ Results are averaged over 5 distinct seeds with standard deviations. The $\widehat{\text{CaCE}}$ value of changing concept $C$ from $c$ to $c'$ is the negative $\widehat{\text{CaCE}}$ value of changing concept $C$ from $c'$ to $c$.}
    \label{tab:cace-results}
    
    \vspace{3pt}
    
    \centering
    
\begin{tabular}{lccc}
\toprule
&           Negative to Positive &            Negative to unknown &            Positive to unknown \\
\midrule
food     &  1.90 (± 0.03) &  1.00 (± 0.02) &  $-$0.82 (± 0.01) \\
service  &  1.42 (± 0.04) &  0.89 (± 0.04) & $-$0.45 (± 0.01) \\
ambiance &  1.27 (± 0.01) &  0.79 (± 0.01) &  $-$0.50 (± 0.03) \\
noise    &  0.75 (± 0.02) &  0.44 (± 0.00) &   $-$0.23 (± 0.02) \\
\bottomrule
\end{tabular}

\end{table}
\egroup

\begin{figure}[tp]
\centering
\includegraphics[width=1\textwidth]{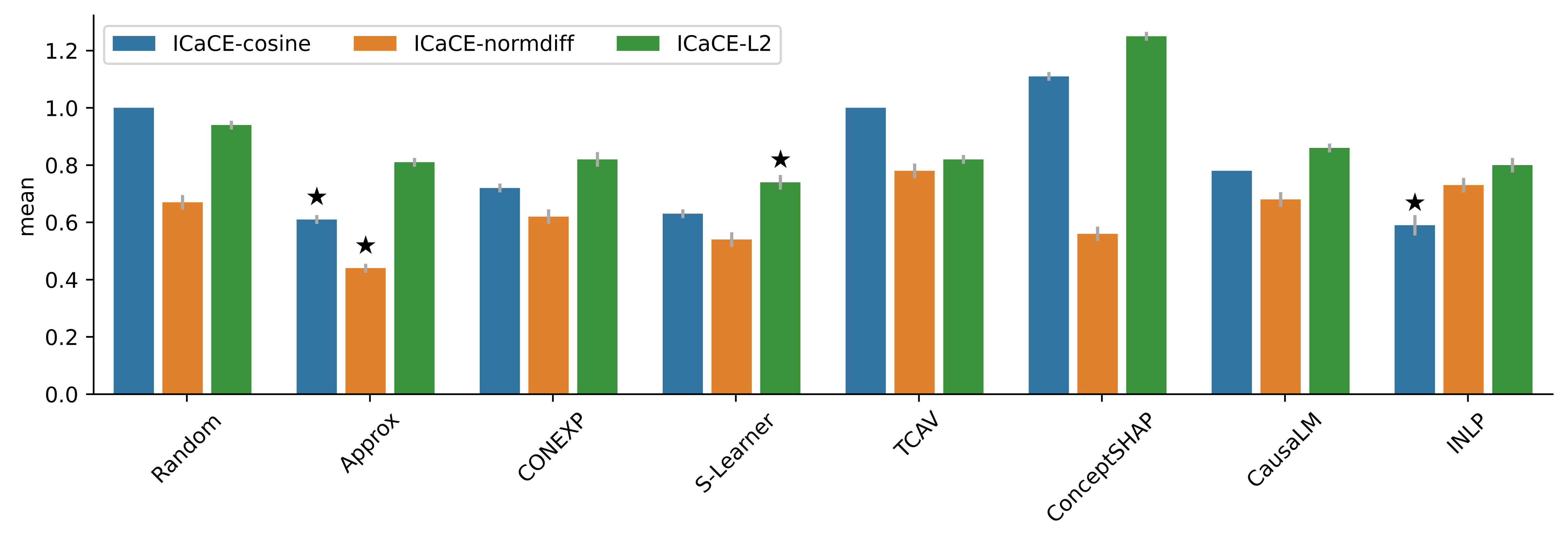}
\caption{$\ICACEerror$ (Definition~\ref{def:ICaCE-Error}) for \texttt{bert-base-uncased} fine-tuned for five-way sentiment, averaged per aspect. We report values for cosine, L2, and normdiff.  \textbf{Lower is better}. Stars mark the best result(s) per metric. Results averaged over 5 distinct seeds. $^{\dagger}$RandomExplainer takes the difference between two random probability vectors as the predicted effect.}
\label{fig:cebab-distance}
\end{figure}

To evaluate the intrinsic capacity of a model to capture causal effects, we report the $\widehat{\text{CaCE}}$ values, as in Definition~\ref{def:CaCE}. The results for \texttt{bert-base-uncased} are given in Table~\ref{tab:cace-results}. They are intuitive and well-aligned with the $\widehat{\text{ATE}}$ estimates in Table~\ref{tab:cebab-ate}, indicating that the model has captured the real-world effects. \footnotetext{Definition~\ref{def:CaCE} defines the CaCE values as vectors. In this table, we collapse the CaCE values to scalars by having $\nn$ output the most probable predicted class, instead of the class distribution.}

Our primary assessment of the evaluation methods is given in Figure~\ref{fig:cebab-distance}, again focusing on a five-way \texttt{bert-base-uncased} model as representative of our results. We provide values based on \textit{cosine}, \textit{L2}, and \textit{normdiff} as the value of $\textrm{Dist}$ in Definition~\ref{eq:ICaCE-Error}. The \textit{cosine}-distance metric measures if the estimated and observed effect have the same direction but does not take the magnitudes of the effects into account. The \textit{L2}-distance measures the Euclidian norm of the difference of the observed and estimated effect. Both the direction and magnitude of the effects influence this metric. To only compare the magnitudes, we use the \textit{normdiff}-distance, which computes the absolute difference between the Euclidean norms of the observed and estimated effects, thus completely ignoring the directions of both effects.

Remarkably, our approximate counterfactual baseline proves to be the best method at capturing both the direction and magnitude of the effects. The fact that a simple baseline method beats almost all other methods indicates that we need better explanation methods if we are going to capture even relatively simple causal effects like those given by \cebab.

Recall from Table \ref{tab:methods-summary} that the compared methods require different levels of access to concept labels at inference time.  Approximate counterfactuals and S-Learner have access to both the direction of the intervention and the predicted test-time aspect labels, enabling them to outperform CONEXP, which has access to only the direction of the intervention, and TCAV, ConceptSHAP, and CausaLM, which have access to neither the intervention direction nor test-time aspect labels.

The INLP method ties with the best method for the \textit{cosine} metric, despite having access to neither intervention directions nor test-time aspect labels. Perhaps this method could be extended to make use of this additional information and decisively improve upon our approximate counterfactual baseline.

While CausaLM and INLP both estimate the effect of removing a concept from an input, INLP uses linear probes to guide interventions on the original model, while CausaLM trains an entirely new model with an auxiliary adversarial objective. The direct use of the original model is something INLP shares with the approximate counterfactual baseline; it seems that a tight connection to the original model may underlie success on \cebab.

\section{Conclusion}\label{sec:conc}

Our main contributions in this paper are twofold. First, we introduced \cebab{}, the first benchmark dataset to support comparing different explanation methods against a single ground-truth with human-created counterfactual texts and multiply-validated concept labels for aspect-level and overall sentiment. Using this resource, one can isolate the true causal concept effect of aspect-level sentiment on any trained overall sentiment classifier. \cebab\ provides a level playing field on which we can compare a variety of explanation methods that differ in their assumptions about their access to the model, their computational demands, their access to ground-truth concept labels at inference time, and their overall conception of the explanation problem. Furthermore, the evaluated methods make absolutely no use of \cebab's counterfactual train set. In turn, we hope that \cebab\ will facilitate the development of explanation methods that can take advantage of the very rich counterfactual structure \cebab\ provides across all its splits.

Second, we have provided an in-depth experimental analysis of how well multiple model explanation methods are able to capture the true concept effect. A naive baseline that approximates counterfactuals through sampling achieves the best performance, with INLP and S-Learner being the only other methods that achieves state-of-the art on any metric. While \cebab\ is only grounded in one task, sentiment analysis alone is enough to produce starkly negative results that should serve as a call to action for NLP researchers aiming to explain their models.

\begin{ack}
This research is supported in part by a grant from Meta~AI.
Karel D'Oosterlinck was supported through a doctoral fellowship from the Special Research Fund (BOF) of Ghent University. We thank our crowdworkers for their invaluable contributions to \cebab.
\end{ack}

\bibliography{custom}

\appendix

\newpage
\clearpage

\section*{Supplementary Materials}

\section{Causal Concept Effects and Metrics for Explanation Methods}\label{app:theory}

Data do not materialize out of thin air. Rather, data are generated from real-world processes with complex causal structures we do not observe directly. Causal inference is the task of estimating theoretical causal effect quantities.

When estimating causal effects, researchers commonly measure the \textit{average treatment effect}, which is the difference in mean outcomes between the treatment and control groups \citep{rubin1974estimating}. Formally, we define the average treatment effect of binary treatment $T$ on an outcome $Y$ under a data generation process $\mathcal{G}$ that represents the unknown details of the real-world.
\begin{definition}[Average Treatment Effect; ATE \citep{rubin1974estimating, pearl1995causal}]
    \begin{equation}\label{eq:ATE-app}
        \ATE_{T}(Y, \mathcal{G}) = \mathbb{E}_\mathcal{G}\big[Y\,\big|\,\doop(T=1)\big] - \mathbb{E}_\mathcal{G}\big[Y \,\big|\,\doop(T=0)\big].
    \end{equation}
\end{definition}

The ATE is a theoretical quantity we cannot compute in practice, since we do not have access to $\mathcal{G}$ nor can we observe both interventions for the same subject.

However, we are concerned with estimating the causal effect of variables representing \textit{non-binary concepts} in real-world systems, on data in an appropriate format for processing by a modern AI model that predicts \textit{vector encoding probability distributions} over outputs. 

Let $\nn $ be a neural network outputting a probability vector, where its $k$-th entry represents the probability to predict the $k$-th class, and let $\phi$ be a feature representation (e.g., \texttt{BERT} embedding). In the context of model explanations, we will define the tools needed to answer three questions:
\begin{enumerate}
\item Given a real-world circumstance $u$ that led to input data $x_u^{C=c}$, what is expected effect of a concept $C$ changing from value $c$ to value $c'$ on the model output of $\mathcal{N}_{\phi}$ provided input data $x_u^{C=c}$?
\item What is the expected effect of a concept $C$ changing from value $c$ to value $c'$ on the output of the model $\mathcal{N}_{\phi}$ provided input data $X$ across real-world circumstances $U$?
\item What is the magnitude of the expected effect of a changing the concept $C$ on  the output of the model $\mathcal{N}_{\phi}$ provided input data $X$ across real-world settings $U$?
\end{enumerate}
For example, in the context of \cebab, we might ask
\begin{enumerate}
\item  Given a real-world dining experience $u$ with good food quality ($C_{\text{food}} = +$) that led to a restaurant review $x^{C_{\text{food}} = +}_u$, what is the effect of changing the food quality $C_{\text{food}}$ from $C_{\text{food}}= +$ to $C_{\text{food}} = -$ on the output of an overall-sentiment text classifier $\mathcal{N}_{\phi}$ provided a review of the dining experience?
\item 
 What is the expected effect of changing the food quality $C_{\text{food}}$ from positive $+$ to negative $-$ on the output of the model $\mathcal{N}_{\phi}$ across real-world dining experiences that lead to restaurant reviews?
\item What is the magnitude of the expected effect of a changing food quality $C_{\text{food}}$ on the output of the model $\mathcal{N}_{\phi}$ across real-world dining experiences that lead to restaurant reviews?
\end{enumerate}
Each of the above questions requires the estimation of a different theoretical quantity. In respect to the order of the questions, these quantities are the \textit{individual causal concept effect}, the \textit{causal concept effect}, and the \textit{absolute causal concept effect}. 

We believe the most practical question in explainable AI is: why does this model have this output behavior for an \textit{actual} input. For this reason, our focus in the main text is \textit{individual causal concept effects}. We define our central metric that captures the performance of an explainer on \cebab\ as the average error on individual causal effect predictions (Definition~\ref{def:ICaCE-Error}). 

We do not evaluate the ability of explainers to evaluate the causal concept effect or the absolute causal concept effect. 

\subsection{Theoretical Quantities} 
\begin{definition}[Causal Concept Effects; \citep{goyal_explaining_2020}]\label{def:cace-app}
For an exogenous setting $u$ that led to concept $C$ taking on value $c$ and the creation of input data $x_u^{C=c}$, the \emph{individual causal concept effect} of a concept $C$ changing from value $c$ to $c'$ in a data generation process $\mathcal{G}$ on a neural network $\nn$ with feature representation $\phi$ is

    \begin{equation}
        \text{ICaCE}_{\mathcal{N}_{\phi}}(\mathcal{G}, x_u^{C=c}, c') =
        \mathbb{E}_{x \sim \mathcal{G}}\left[ \nn \big(\phi(x)\big) \, \big| \, \doop\left(       
        \begin{array}{@{} c @{}}
        C=c' \\
        U=u 
        \end{array}
        \right) 
    \right] - \nn \big(\phi(x^{C=c}_u)\big) 
    \end{equation}
    
The \emph{causal concept effect} is the effect in general, meaning there is no input data generated from a fixed exogenous real-world setting:
 \begin{equation}
     \text{CaCE}_{\mathcal{N}_{\phi}}(\mathcal{G}, C, c, c') = 
        \mathbb{E}_{x \sim \mathcal{G}}\big[ \nn \big(\phi(x)\big) \, \big| \, do(
        C=c' 
        ) \big] -   \mathbb{E}_{x \sim \mathcal{G}}\big[ \nn \big(\phi(x)\big) \, \big| \, do(
        C=c 
        ) \big]
 \end{equation}
The \emph{absolute causal concept effect} estimate of the magnitude of the effect a concept has on a classifier output, regardless the concept values. We aggregate over all possible intervention values in the following way
\begin{equation}%
 \text{ACaCE}_{\mathcal{N}_{\phi}}(\mathcal{G}, C) = \frac{1}{|\{\{ c, c' \} \subseteq {C}\}|}\sum_{\{ c, c' \} \subseteq {C}} \left| \text{CaCE}_{\mathcal{N}_{\phi}}(\mathcal{G}, C, c, c') \right|,
\end{equation}
where ${C}$ is the set of all possible values for concept in addition to denoting the concept itself.\footnote{We take the absolute value since $\text{CaCE}_{\mathcal{N}_{\phi}}(\mathcal{G}, C, c, c') = -\text{CaCE}_{\mathcal{N}_{\phi}}(\mathcal{G}, C, c', c)$, and these cancel each other in the summation.}
\end{definition}

\subsection{Empirical Estimates} 

Similar to the ATE, causal concept effects are theoretical quantities we can only estimate in reality. To perform such estimates, we need a dataset consisting of pairs $(x^c_u, x^{c'}_u) \in \mathcal{D}$ that are drawn from a data generation process $\mathcal{G}$. A major contribution of this work is crowdsourcing such a dataset, \cebab. These pairs allow us to compute empirical estimations of (individual) causal concept effects.

\begin{definition}[Empirical Causal Concept Effects]\label{def:empirical-cace-app}
For an exogenous setting $u$, the empirical individual causal concept effect of a concept $C$ changed from value $c$ to $c'$, for $\mathcal{D}$ sampled from $\mathcal{G}$, on a neural network $\nn $ trained on a feature representation $\phi$ is
    \begin{equation}
        \widehat{\text{ICaCE}}_{\mathcal{N}_{\phi}}(x_u^{C=c'}, x_u^{C=c}) =  \nn \big(\phi(x^{C=c'}_u)\big) - \nn \big(\phi(x^{C=c}_u)\big)
    \end{equation}
Given a full dataset $\mathcal{D}$ of such pairs, we can estimate the causal concept effect
\begin{equation}
    \widehat{\text{CaCE}}_{\mathcal{N}_{\phi}}(\mathcal{D}, C, c, c') = \frac{1}{|\mathcal{D}_{C}^{c \to c'}|} \sum_{(x^c_u, x^{c'}_u) \in \mathcal{D}}  \widehat{\text{ICaCE}}_{\mathcal{N}_{\phi}}(x_u^{C=c}, x_u^{C=c'})
\end{equation}
And also the absolute causal concept effect
\begin{equation}
    \widehat{\text{ACaCE}}_{\mathcal{N}_{\phi}}(\mathcal{D}) = \frac{1}{|\{\{ c, c' \} \subseteq {C}\}|} \sum_{(c, c') \in {C}} \lvert      \widehat{\text{CaCE}}_{\mathcal{N}_{\phi}}(\mathcal{D},C,c,c')\rvert
\end{equation}
\end{definition}

Notice that the only difference between causal concept effects (Definition~\ref{def:cace-app}) and empirical causal concept effects (Definition~\ref{def:empirical-cace-app}) is that we change the expectation taken over $\mathcal{G}$ to be the average over a dataset $\mathcal{D} \sim \mathcal{G}$.

\subsection{Explainer Errors}

Given a dataset $\mathcal{D}$ and an explainer $\mathcal{E}_{\mathcal{N}_{\phi}}(x^c_u,c')$ that predicts individual causal concept effects $\textit{ICACE}_{\mathcal{N}_{\phi}}(x^c_u,c')$, we define metrics capturing the ability of $\mathcal{E}$ to estimate causal effects by simple computing the averaged distance between our explainer and the empirical causal effect
\begin{definition}[Explainer Distances]\label{def:cebab-score}
The average distance between the explainer and the empirical individual causal concept effects.
\begin{multline}
    \text{ICaCE-Error}_{\nn_{\phi}}^{\mathcal{D}}(\mathcal{E}, C, c, c') = \\
    \frac{1}{\left|\mathcal{D}_{C}^{c \to c'}\right|} 
    \negthickspace
    \sum_{(x^{C=c}_u, x^{C=c'}_u) \in \mathcal{D}_C^{c \to c'}}
    \negthickspace
    \mathsf{Dist}\big(
    \widehat{\text{ICaCE}}_{\nn_{\phi}}(x^{C=c}_u, x^{C=c'}_u), 
    \mathcal{E}_{\nn_{\phi}}(x^{C=c}_u, x^{C=c'}_u)
    \big)
\end{multline}
The distance between the average of explainer outputs and the empirical causal concept effect
\begin{equation}
\textit{CaCE-Error}_{\nn_{\phi}}^{\mathcal{D}}(\mathcal{E}, C, c, c') =  
    \lVert   
    \widehat{\text{CaCE}}_{\mathcal{N}_{\phi}}(\mathcal{D}, C, c, c'), \; \frac{1}{
    \left|\mathcal{D}_C^{c \to c'}\right|} \sum_{x^c_u, x^{c'}_u \in \mathcal{D}_C^{c \to c'}}  \mathcal{E}_{\mathcal{N}_{\phi}}(x^c_u,c')\big)\rVert 
\end{equation}
The distance between the average magnitude of explainer outputs and the empirical absolute causal effect
\begin{multline}
\textit{ACaCE-Error}_{\nn_{\phi}}^{\mathcal{D}}(\mathcal{E}, C) =\\    
\lVert    
\widehat{\text{ACaCE}}_{\mathcal{N}_{\phi}}(\mathcal{D}, C), \;\; \frac{1}{|\{\{ c, c' \} \subseteq {C}\}|} \sum_{(c, c') \in {C}} \frac{1}{\left|\mathcal{D}^{c \to c'}_C\right|} \sum_{x^c_u, x^{c'}_u \in \mathcal{D}^{c \to c'}_C}  |\mathcal{E}_{\mathcal{N}_{\phi}}(x^c_u,c')\big)| \rVert
\end{multline}

where $\| \cdot \|$ is some distance metric and $\mathcal{D}_{C}$ is the subset of data where $C$ is the concept changed and $\mathcal{D}_C^{c \to c'}$ is the subset of data where $C$ is the concept changed from value $c$ to value $c'$.
\end{definition}

In the main text, we use the ICaCE-Error as our primary evaluation metric.

\section{\cebab}\label{app:cebab}

Our supplementary materials contain a full Datasheet for \cebab\ as a separate markdown document.

\subsection{Restaurant-level metadata from \opentable{}}

Table~\ref{tab:cebab-meta} gives an overview of the metadata associated with the original review texts in \cebab{}. 

\begin{table}[!ht]
\caption{\cebab\ metadata from \opentable, tabulated at the level of individual original reviews. A total of 1,084 restaurants are represented in the data.}
  \label{tab:cebab-meta}
  
  \vspace{3pt}
  
  \centering
  \setlength{\tabcolsep}{4pt}
  \begin{subtable}[b]{0.3\linewidth}
    \centering    
    \begin{tabular}{lr}
      \toprule
      italian       &     1076 \\
      american      &      654 \\
      french        &      254 \\
      seafood       &      202 \\
      mediterranean &      113 \\
      \bottomrule
    \end{tabular}
    \caption{Cuisine.}
    \label{tab:cebab-meta-cuisine}
  \end{subtable}
  \hfill
  \begin{subtable}[b]{0.3\linewidth}
    \centering
    \begin{tabular}{lr}
      \toprule     
      northeast &     863 \\
      west      &     634 \\
      south     &     470 \\
      midwest   &     332 \\
      \bottomrule
    \end{tabular}
    \caption{U.S.~regions.}
    \label{tab:cebab-meta-region}
  \end{subtable}
  \hfill
  \begin{subtable}[b]{0.3\linewidth}
    \centering
    \begin{tabular}{lr}
      \toprule
      1 star &             244 \\
      2 star &            1207 \\
      3 star &             123 \\
      4 star &             330 \\
      5 star &             395 \\
      \bottomrule
    \end{tabular}
    \caption{Star ratings.}
    \label{tab:cebab-meta-stars}
  \end{subtable}    
\end{table}

\subsection{Crowdworkers}

A total of 254 workers participated in our experiments. All of them come from a pool of workers whom we prequalified to participate in our tasks based on the work they did for us on previous crowdsourcing projects. Thus, we expected that they would do high quality work, and they more than lived up to our expectations, as indicated by the high degree of success they achieved when editing and the high degree of consensus they reached about how to label examples.

There are a total of 642 instances of 15,0006 for which, despite our best efforts, a worker validated an example that they themselves created during the editing phase. Removing the contributions of these workers affects the majority in only 24 cases, with no clear pattern to the changes, so we kept all the validation labels in order to ensure that every example has give responses.

\subsection{Editing Phase}

\begin{figure}[tp]
  \centering
  \includegraphics[width=0.8\textwidth]{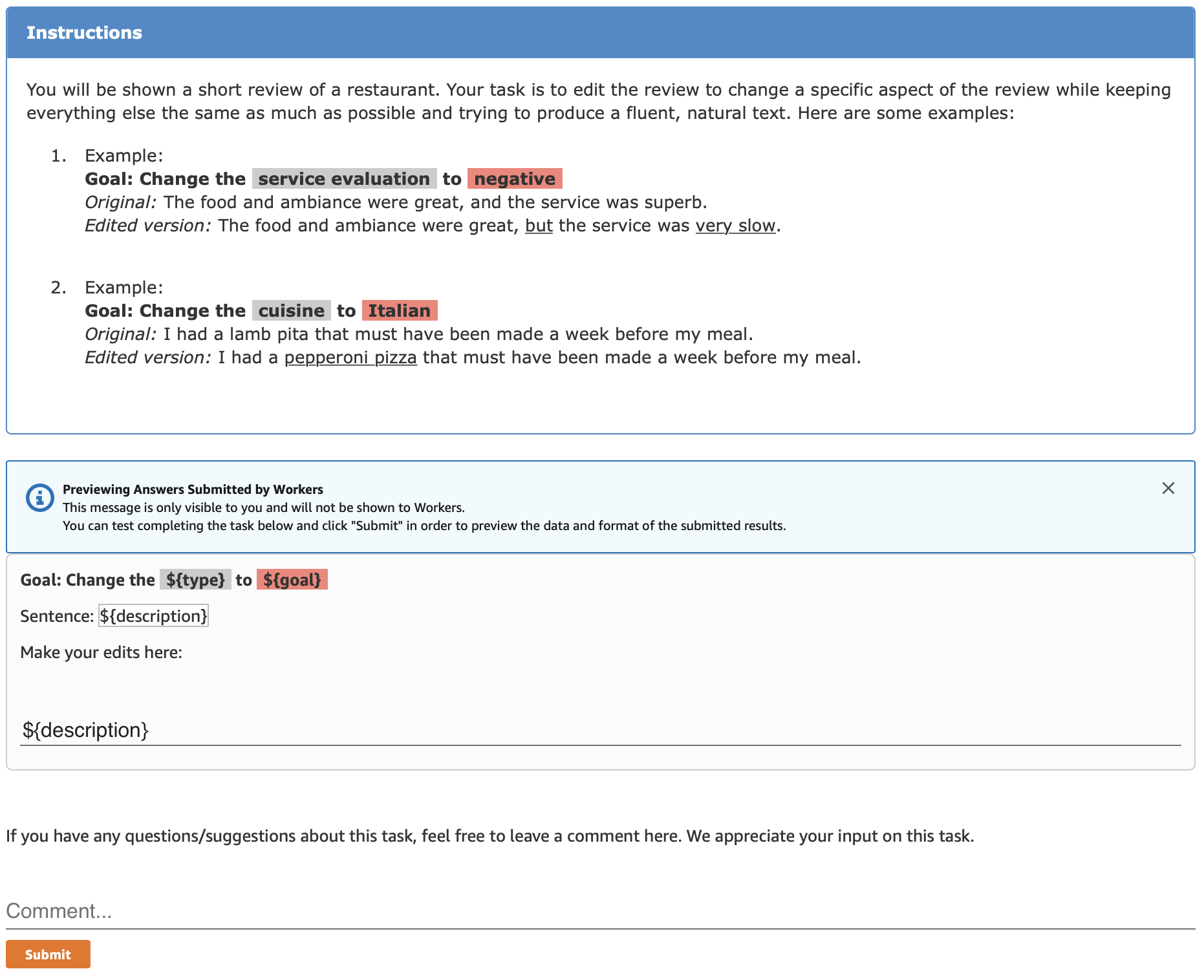}
  \caption{Edit phase annotation interface where the task was to convey `Positive' or `Negative' for the target aspect.}
  \label{fig:mturk-edit}
\end{figure}

\begin{figure}[tp]
  \centering
  \includegraphics[width=0.8\textwidth]{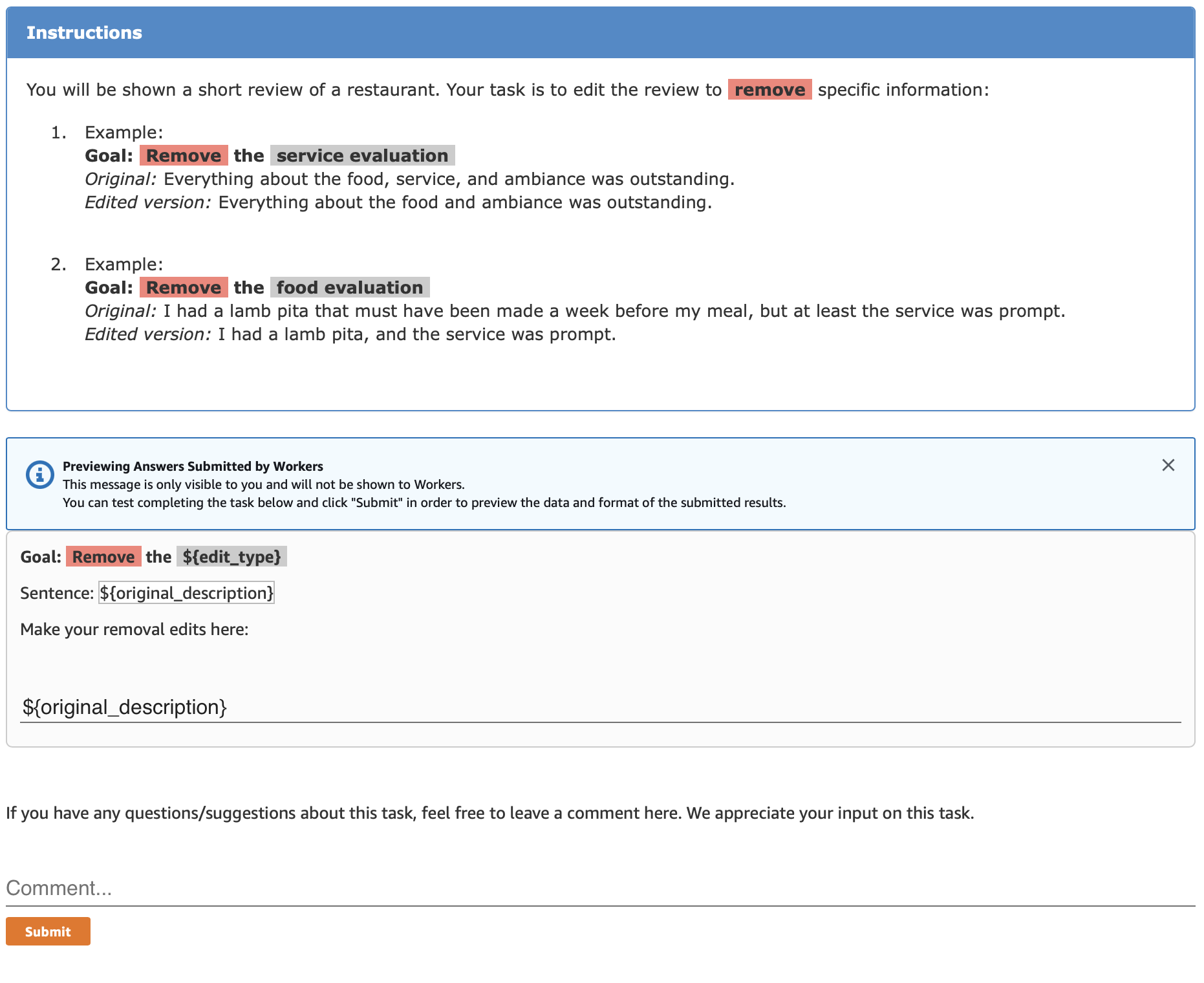}
  \caption{Edit phase annotation interface where the task was to hide the sentiment of the target aspect.}
  \label{fig:mturk-unk}
\end{figure}

A total of 183 workers participated in this phase. Workers were paid US\$0.25 per example. Figure~\ref{fig:mturk-edit} shows the annotation interface that workers used when changing the target aspect's sentiment to either `Positive' or `Negative', and Figure~\ref{fig:mturk-unk} shows the interface where the task was to hide the target aspect's sentiment.

\begin{figure}[ht]
  \centering
  \includegraphics[width=0.6\textwidth]{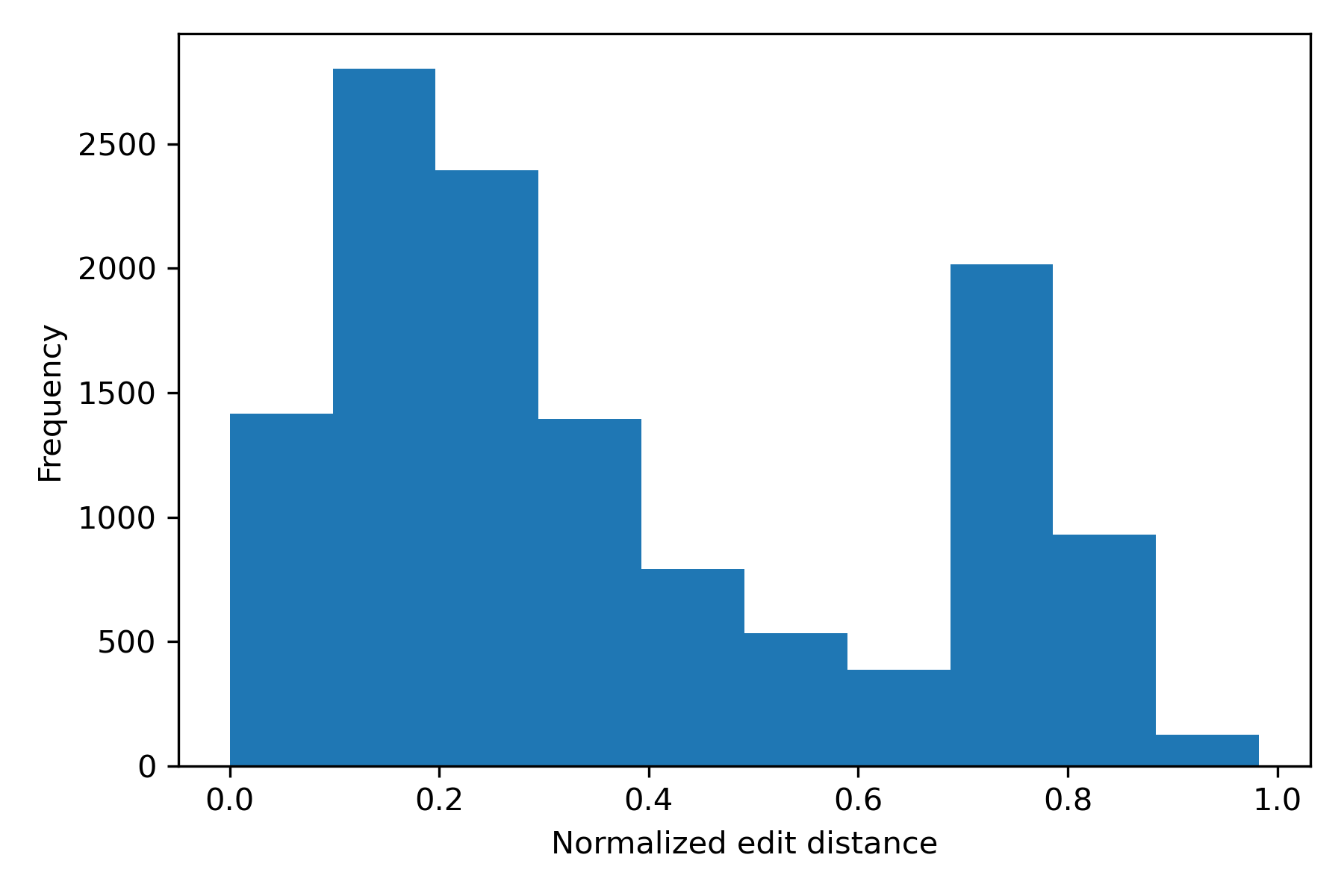}
  \caption{Normalized edit distances between original texts and those created during the editing phase for \cebab.}
  \label{fig:edit-distances}
\end{figure}

Figure~\ref{fig:edit-distances} summarizes the distribution of edit distances between original and edited texts. These distances are calculated at the character-level and normalized by the length of the original or review, whichever is longer.

\subsection{Validation Phase}

\begin{figure}[tp]
  \centering
  \includegraphics[width=0.8\textwidth]{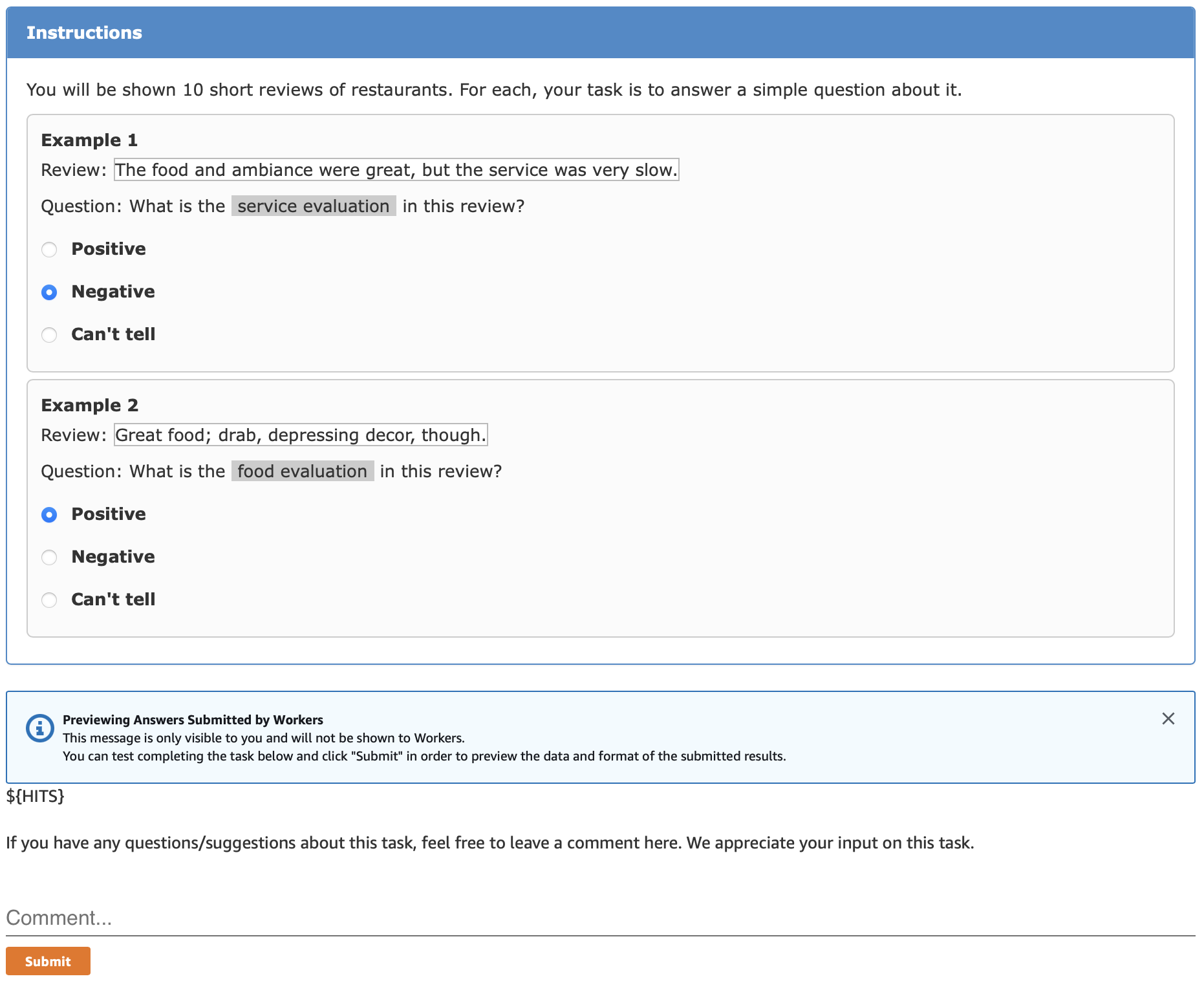}
  \caption{Validation phase annotation interface.}
  \label{fig:mturk-validate}
\end{figure}

A total of 174 workers participated in this phase. Workers were paid US\$0.35 per batch of 10 examples. Figure~\ref{fig:mturk-validate} shows the annotation interface that workers used.

\subsection{Review-level Rating Phase}

\begin{figure}[tp]
  \centering
  \includegraphics[width=0.8\textwidth]{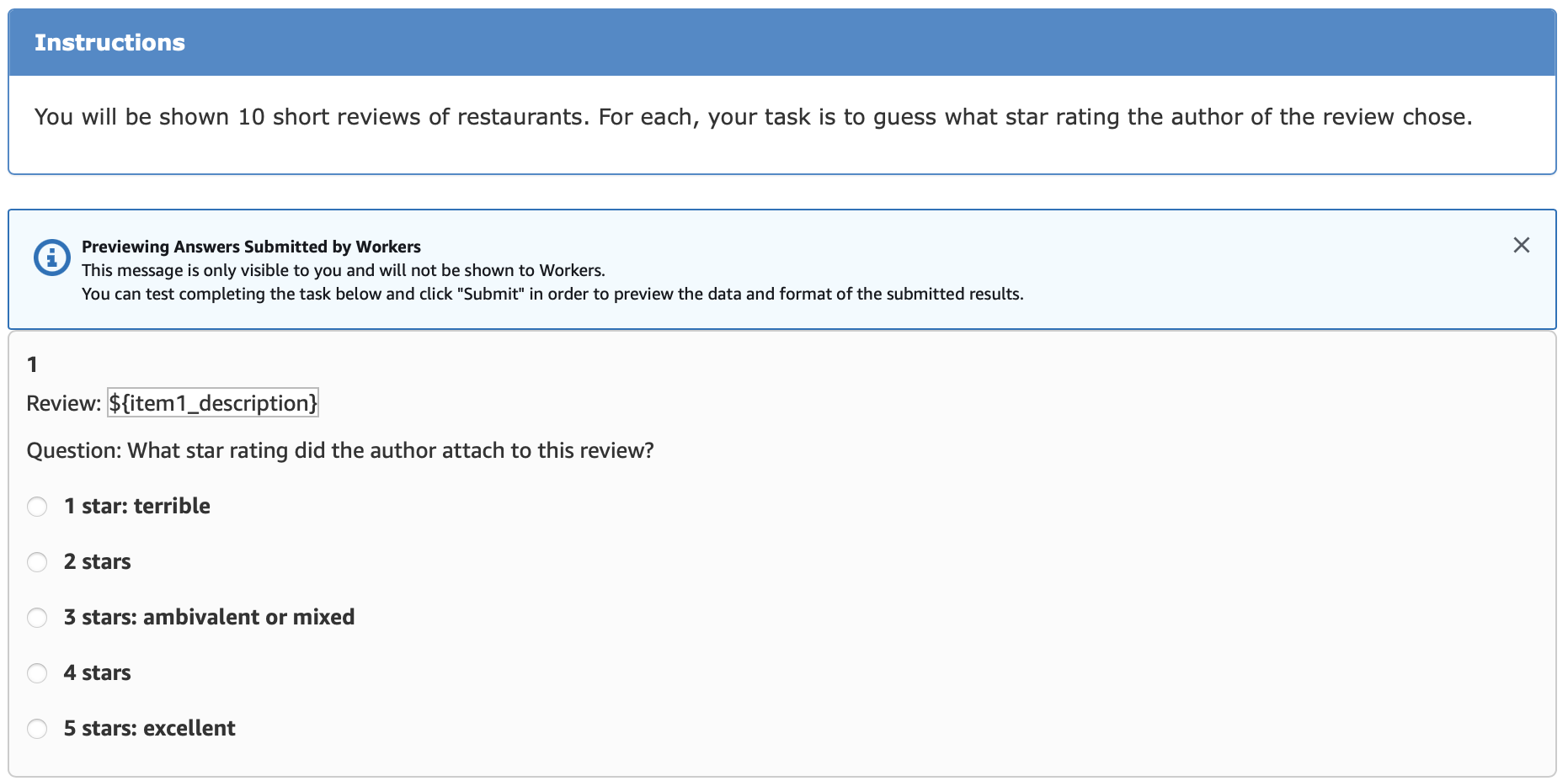}
  \caption{Review-level annotation interface.}
  \label{fig:mturk-review}
\end{figure}

A total of 155 workers participated in this phase. Workers were paid US\$0.35 per batch of 10 examples. Figure~\ref{fig:mturk-review} shows the annotation interface that workers used.

\subsection{Randomly Selected Examples}

Table~\ref{tab:cebab-examples} provides a random sample of edit pairs from \cebab{}'s dev set.

\begin{sidewaystable}[tp]
\caption{Randomly sampled edit pairs from \cebab.}
\label{tab:cebab-examples}

\vspace{3pt}

\setlength{\tabcolsep}{2pt}
\small
\centering
\begin{tabular}{p{0.5\textwidth} *{7}{c} }
\toprule
  description &  original? &    aspect & edit goal & aspect labels & aspect maj. & review labels & review maj. \\
\midrule
                                                                                                                     Food was disgusting and very unreasonable!!!!!! Every request was honored and very friendly staff.\textbackslash nHomemade bread which was foul....... &        False &      food &  -- &        --, --, --, --, -- &        -- &             2, 2, 2, 2, 2 &               2 \\
                                                                                                                                                                                                           Every request was honored and very friendly staff. &        False &      food &   unk. &       unk, unk, unk, +, + &         unk. &             5, 5, 5, 4, 4 &               5 \\\midrule
                                                                                                                                                                                                          The food was average, but the service was terrible. &         True &      food &      None &        --, --, --, unk, + &        -- &             2, 2, 2, 3, 3 &               2 \\
                                                                                                                                                                                                    The food was above average, but the service was terrible. &        False &      food &  + &             +, +, +, +, + &        + &             3, 3, 3, 3, 2 &               3 \\\midrule
                                                                                                                                                                                                                        We hated our afternoon at Shorebreak! &        False &  ambiance &  -- &      --, --, --, unk, unk &        -- &             1, 1, 1, 1, 1 &               1 \\
                                                                                                                                                                                                                        We loved our afternoon at Shorebreak! &        False &  ambiance &   unk. &      unk, unk, unk, --, + &         unk. &             5, 5, 5, 4, 4 &               5 \\\midrule
 The Sunday Jazz Brunch is great - Good music and fine, creative food. The service was great, my server answered all of my questions. The ambiance is quiet, but not so quiet as to inhibit conversation. A wonderful way to spend an early Sunday afternoon. &        False &   service &  + &             +, +, +, +, + &        + &             5, 5, 5, 4, 4 &               5 \\
                      The Sunday Jazz Brunch is great - Good music and fine, creative food. The ambiance is quite, but not so quite as to inhibit conversation. A wonderful way to spend an early Sunday afternoon. The only bad spot was the horrid service. &        False &   service &  -- &        --, --, --, --, -- &        -- &             4, 4, 4, 4, 3 &               4 \\\midrule
                                                                                                                                     My pasta dish was flavorless and rubbery and my husband's was cold. At least it 45 minutes to get it. Very poor, indeed. &         True &      food &      None &        --, --, --, --, -- &        -- &             1, 1, 1, 2, 2 &               1 \\
                                                                                                                                                             My pasta dish was amazing and cooked great. At least it 45 minutes to get it. Very poor, indeed. &        False &      food &  + &            +, +, +, +, -- &        + &             3, 3, 3, 3, 1 &               3 \\\midrule
                                                                                                                                                                                      liked the restaurant a lot and loved the meal. Found the chicken great! &        False &      food &  + &             +, +, +, +, + &        + &             5, 5, 5, 4, 3 &               5 \\
                                                                                                                                                                                                                                I liked the restaurant a lot, &        False &      food &   unk. &     unk, unk, unk, unk, + &         unk. &             5, 5, 5, 4, 4 &               5 \\\midrule
                                                                                                                                                                                                              At the heart of it, this is a HOTEL restaurant. &         True &     noise &      None &   unk, unk, unk, unk, unk &         unk. &             3, 3, 3, 3, 2 &               3 \\
                                                                                                                                                                                                    At the heart of it, this is an extremely loud restaurant. &        False &     noise &  -- &        --, --, --, --, -- &        -- &             1, 1, 1, 3, 2 &               1 \\\midrule
                                                       I was expecting some dishes from the Northern Italian Cuisine. The menu was not distinguishable from any other chain. The food was good but no differentiation. It was noisy, but I believe by design. &         True &      food &      None &             +, +, +, +, + &        + &             3, 3, 3, 4, 2 &               3 \\
                                                         I was expecting some dishes from the Northern Italian Cuisine. The menu was not distinguishable from any other chain. The food was even worse than that. It was also noisy, but I believe by design. &        False &      food &  -- &         --, --, --, --, + &        -- &             1, 1, 1, 2, 2 &               1 \\
\bottomrule
\end{tabular}
\end{sidewaystable}

\subsection{Five-way Empirical ATE for \cebab}

Table~\ref{tab:cebab-ate-binary} provides the binary $\widehat{\text{ATE}}$ values for \cebab. These can be compared with the corresponding five-way values in Table~\ref{tab:cebab-ate} in the main text.

\begin{table}[tp]
 \caption{Empirical $\widehat{\text{ATE}}$ for the binary sentiment labels in \cebab. Reversing concept order results in the negation of the value given.}
    \label{tab:cebab-ate-binary}
    
    \vspace{3pt}
    
    \centering
     \begin{tabular}{l rrr}
    \toprule
     & Neg to Pos & Neg to Unk & Pos to Unk \\
    \midrule
    food      &  0.77 &  0.49 & $-$0.41 \\
    service   &  0.25 &  0.20 & $-$0.16 \\
    ambiance  &  0.14 &  0.18 & $-$0.14 \\
    noise     &  0.08 &  0.04 & $-$0.14 \\
    \bottomrule
    \end{tabular}
\end{table}

\subsection{Edit variability}

In the editing phase we ask human annotators to produce edits of an original review with regard to some concept. This is inherently a noisy process, which may impact the quality of our final benchmark. The \cebab~dataset features a modest set of paired edits (176 pairs in total). Each of these pairs contains two edits, starting from the same original sentence and edit goal, which results in two different edited sentences. Like all sentences in \cebab, these edits were labeled for their review score by human annotators.

\begin{figure}
    \centering
    \begin{subfigure}{0.48\textwidth}
        \includegraphics[width=\textwidth]{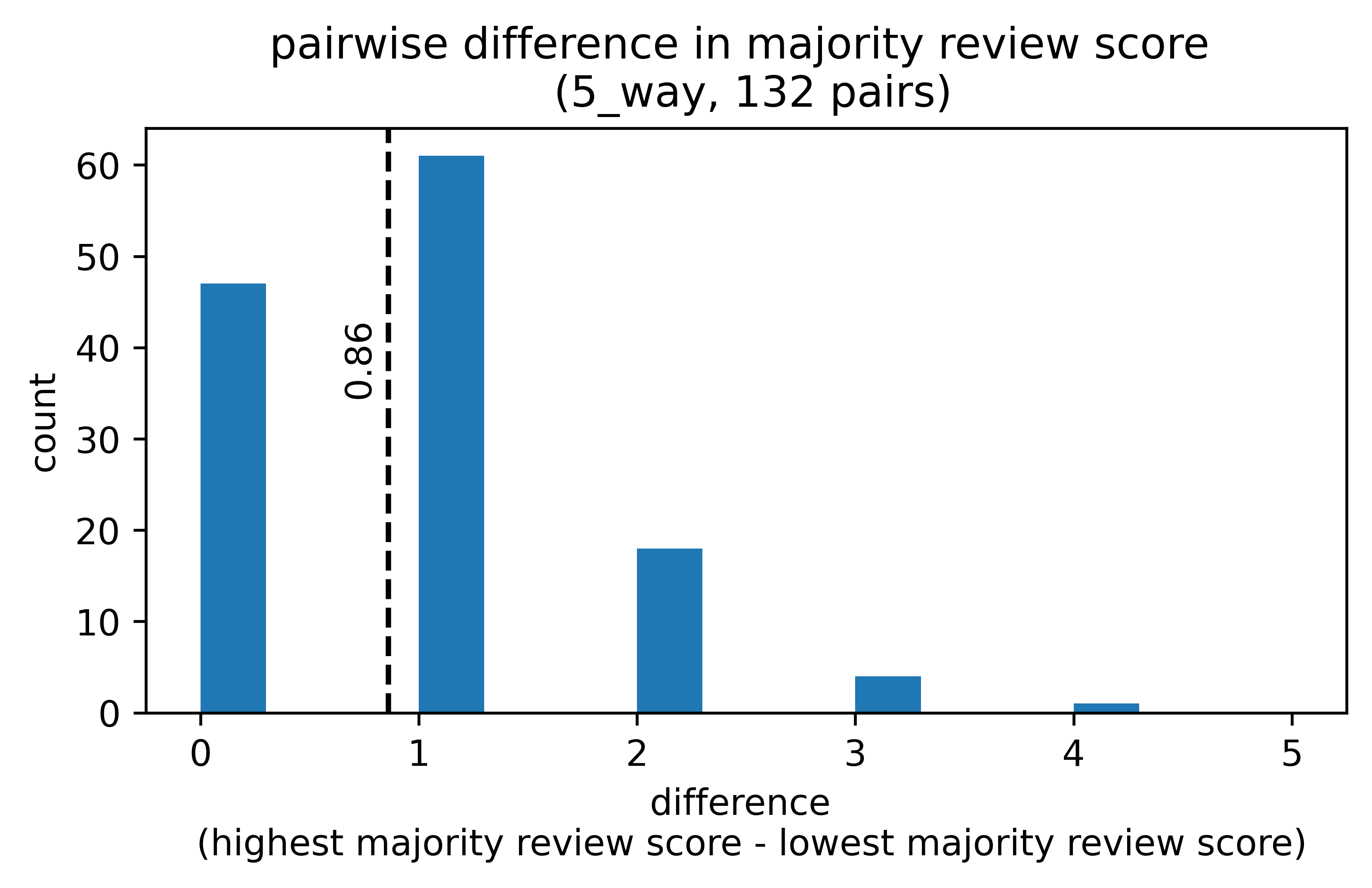}
        \caption{} \label{fig:edit_variability_histplot_a}
    \end{subfigure}
    \hfill
    \begin{subfigure}{0.48\textwidth}
        \includegraphics[width=\textwidth]{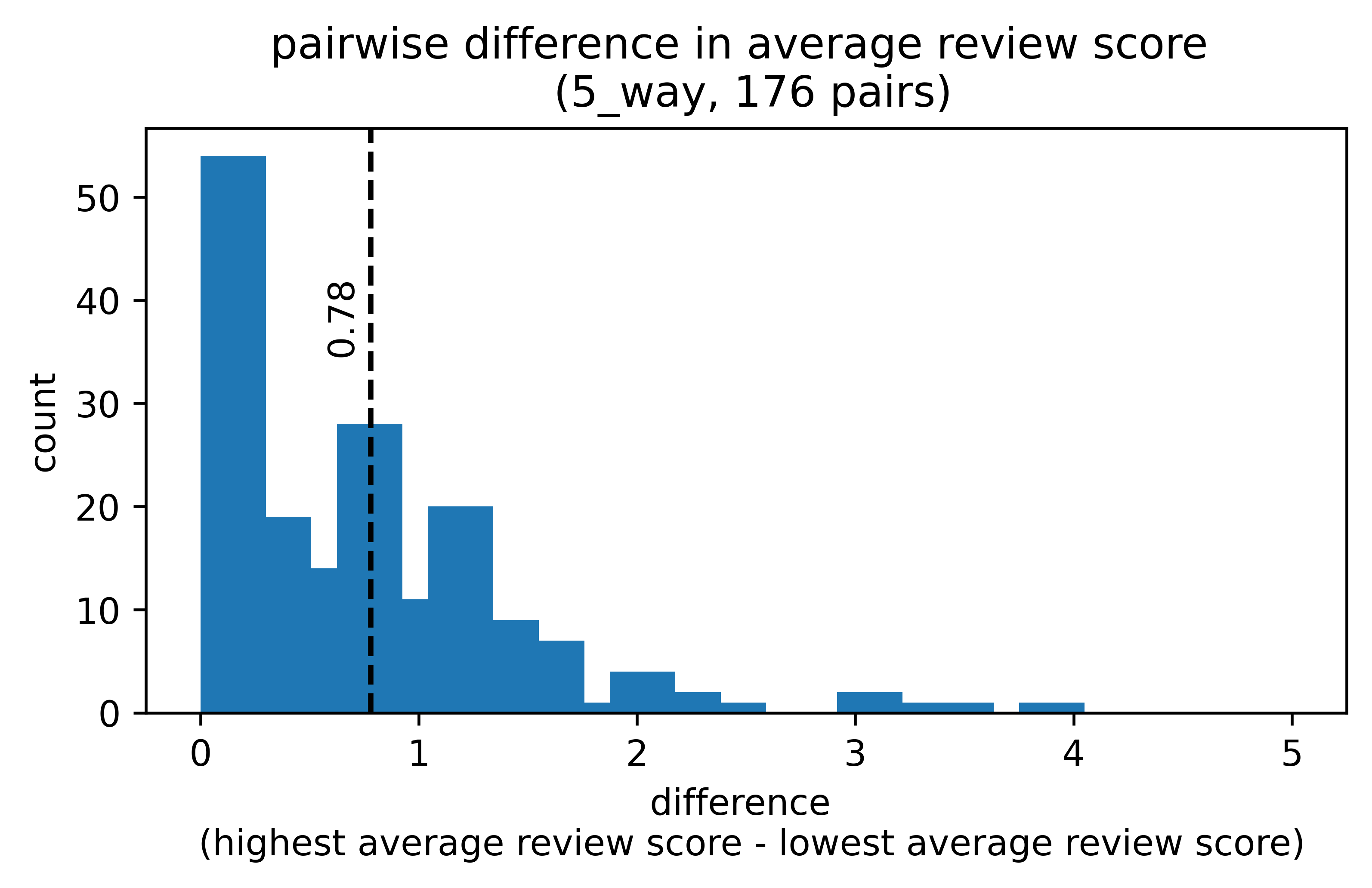}
        \caption{} \label{fig:edit_variability_histplot_b}
    \end{subfigure}
    \caption{Pairwise absolute difference in majority (a) and average (b) review score for all double edits. Figure (a) only considers the 132 pairs where both edits have an actual review majority. Figure (b) considers all 176 pairs. Averages of the distributions are shown with a dotted vertical line.}
    \label{fig:edit_variability_histplot}
\end{figure}

Figure \ref{fig:edit_variability_histplot_a} shows the distribution of the difference in final review majorities produces by these paired edits. Most paired edits differ at most by one star in their final majority rating, indicating that in general there is some noise associated with the editing procedure, but this does not have a major impact on the final review score. Figure \ref{fig:edit_variability_histplot_b} shows the same distribution when we consider the average review score an edit received, as opposed to the majority score. If we consider these average scores, most of the paired edits differ only slightly in their resulting review score.

\begin{figure}
    \centering
    \begin{subfigure}{0.28\textwidth}
        \includegraphics[width=\textwidth]{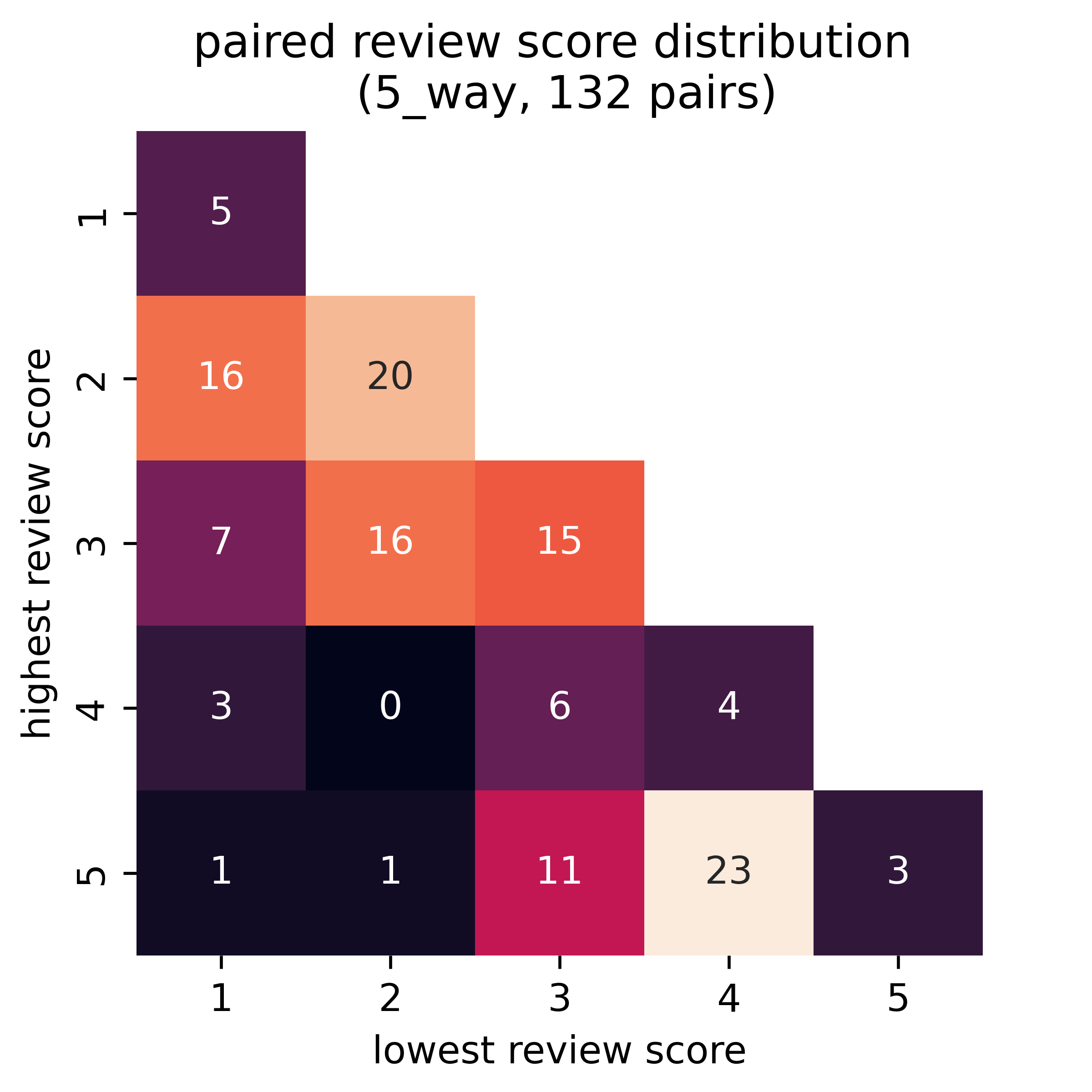}
        \caption{} \label{fig:edit_variability_heatmap_a}
    \end{subfigure}
    \hfill
    \begin{subfigure}{0.28\textwidth}
        \includegraphics[width=\textwidth]{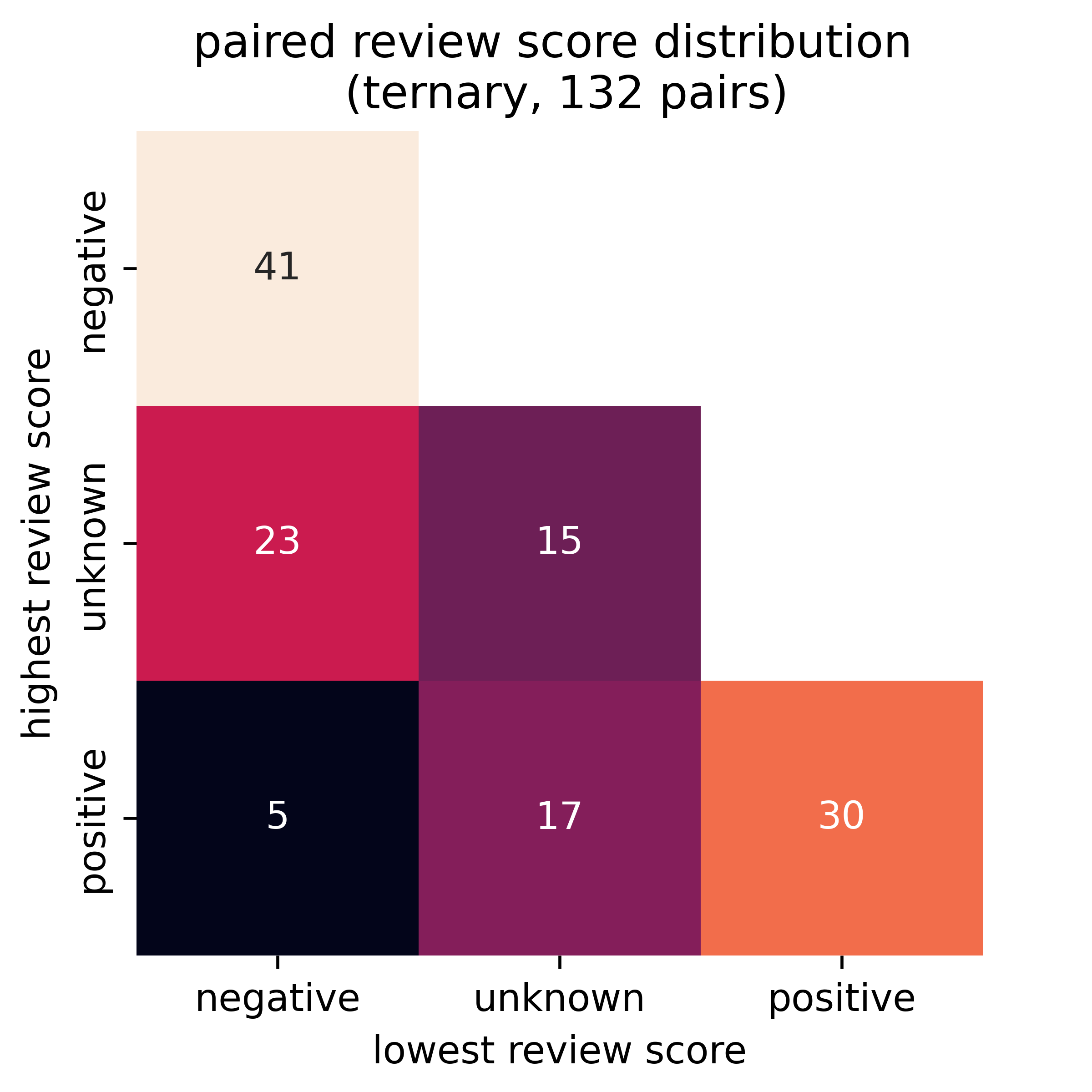}
        \caption{} \label{fig:edit_variability_heatmap_b}
    \end{subfigure}
    \hfill
    \begin{subfigure}{0.28\textwidth}
        \includegraphics[width=\textwidth]{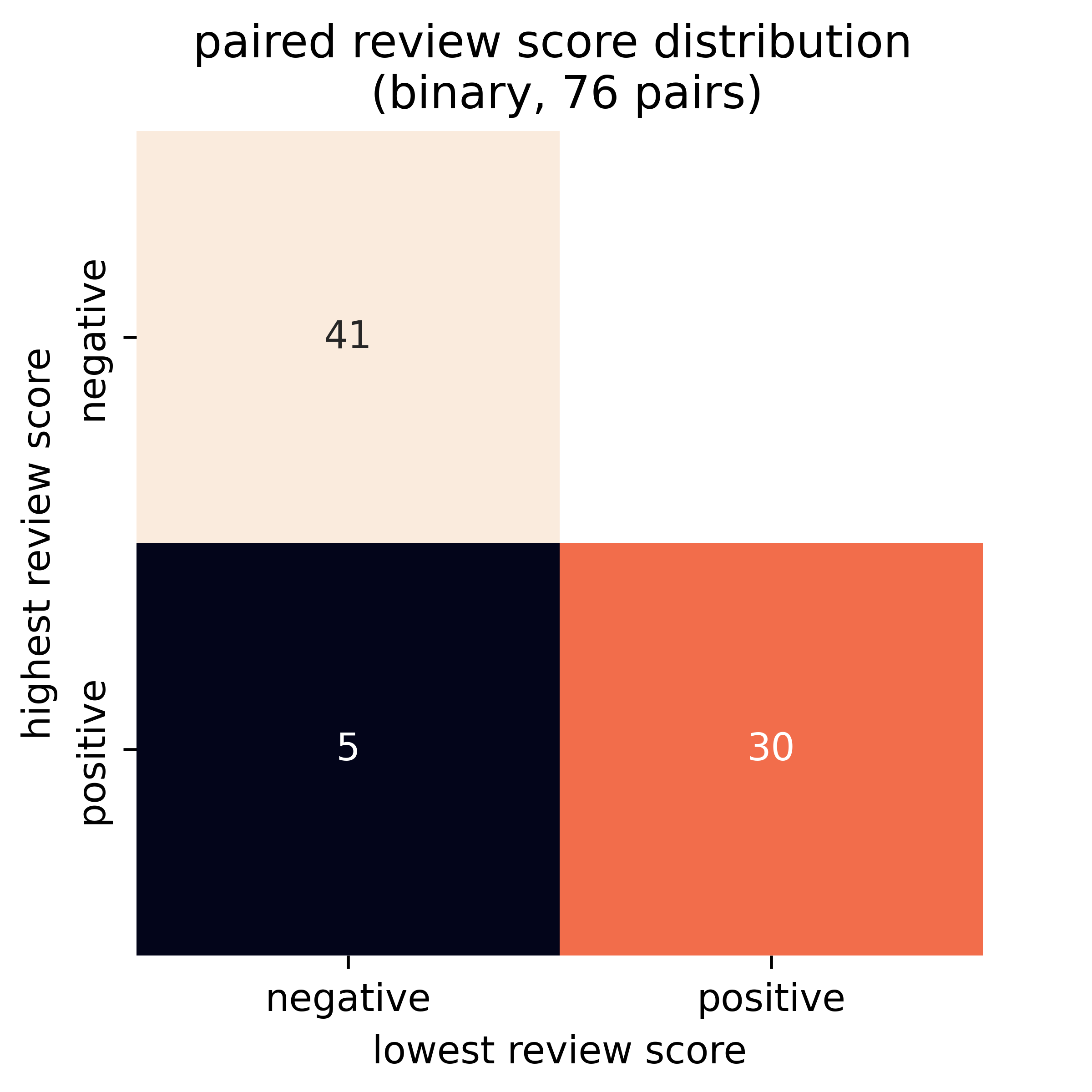}
        \caption{} \label{fig:edit_variability_heatmap_c}
    \end{subfigure}
    \caption{Pairwise review majority distribution for all double edits in 5-way (a), ternary (b), and binary (c) classification settings. Figures (a) and (b) consider only the 132 pairs where both edits have an actual review majority. Figure (c) considers the 76 pairs that have both a review majority and non-neutral labels.}
    \label{fig:edit_variability_heatmap}
\end{figure}

Figures \ref{fig:edit_variability_heatmap}a-c shows the distribution of this pairwise review score in more detail. In an idealized setting without variability, the distribution would be centered around the diagonal of the heatmap. When going from 5-way classification to ternary and binary classification, the variability introduced by the edits becomes less relevant with regard to the final review majority label.

\section{\cebab\ Modeling Experiments}\label{app:cebab-models}

This section reports on standard classifier-based experiments with \cebab, aimed at providing a sense for the dataset when it is used as a standard supervised sentiment dataset. We report experiments on the aspect-level and review-level ratings. In addition, we present evidence that author identity does not have predictive value.

\subsection{Experiments Set-up}

We rely on the Hugging Face \texttt{transformers} library.\footnote{\url{https://github.com/huggingface/transformers}}~\citep{wolf2019huggingface} We train our models with 4 Nvidia 2080 Ti RTX 11GB GPUs on a single node machine. We use a maximum sequence length of 128 with a fix batch size of 32 with a initial learning rate of $2e^{-5}$. We run each experiment 5 times with distinct random seeds. We train our models with a minimum epoch number of 5 with our largest training set. We linearly scale our training epoch number by the size of the training set. We skip hyperparameter tuning for optimized task performance as our goal for this paper is to evaluate explanation methods. We release all of our models on Huggingface Dataset Hub.

\subsection{Models}
We include 4 different types of models, including BERT (\texttt{bert-base-uncased})~\citep{devlin_bert_2019}, RoBERTa (\texttt{roberta-base})~\citep{liu_roberta_2019}, GPT-2 (\texttt{gpt2})~\citep{radford2019language}, %
as well as LSTM with dot-attention~\citep{luong2015effective}. Our LSTM model uses \texttt{bert-base-uncased} tokenizer for simplicity. 
We initialize the embeddings of tokens for our LSTM using fastText~\citep{joulin2016fasttext}. We reconfigure the classification head all other models the same classification head as in RoBERTa as a non-linear multilayer perceptron (MLP).%
\footnote{We implemented T5 (\texttt{t5-base};~\citep{raffel2019exploring}) as a text-to-text model with the goal of treating predicted tokens as class labels. However, this raised unanticipated implementation questions concerning how to post-process multi-token class labels (e.g., ``very positive'') for use in our explainer methods. As a result, we have elected to leave the T5 results out of the current draft, but we intend to include them in the next version once they have been more thoroughly vetted.}

\begin{table*}[tp]
\caption{Model performance results for sequence classification as well as aspect-based sentiment analysis (ABSA) under 3 training conditions. Mean Macro-F1 scores across 5 runs with distinct random seeds are reported.}
  \label{tab:seq-cls-perf}
  
  \vspace{3pt}
  
\centering
\resizebox{1.0\linewidth}{!}{%
  \centering
  \setlength{\tabcolsep}{12pt}
  \begin{tabular}[c]{l c c c c c c c c}
    \toprule
        ~ & \multicolumn{4}{c}{\textbf{Exclusive}} & \multicolumn{4}{c}{\textbf{Inclusive}} \\
        \textbf{Model} & Binary & Ternary & 5-way & ABSA & Binary & Ternary & 5-way & ABSA  \\
    \midrule
    \multicolumn{9}{c}{dev split} \\
    \midrule
        \texttt{BERT} & 0.97 & 0.82 & 0.68 & 0.88 & 0.98 & 0.85 & 0.72 & 0.90 \\
        \texttt{GPT-2} & 0.97 & 0.80 & 0.67 & 0.88 & 0.98 & 0.84 & 0.70 & 0.89 \\
        \texttt{LSTM} & 0.94 & 0.75 & 0.59 & 0.83 & 0.96 & 0.82 & 0.68 & 0.87 \\
        \texttt{RoBERTa} & 0.99 & 0.83 & 0.71 & 0.89 & 0.99 & 0.86 & 0.76 & 0.90 \\ 
    \midrule
    \multicolumn{9}{c}{test split} \\
    \midrule
        \texttt{BERT} & 0.97 & 0.82 & 0.70 & 0.87 & 0.98 & 0.84 & 0.73 & 0.89 \\
        \texttt{GPT-2} & 0.97 & 0.80 & 0.65 & 0.87 & 0.97 & 0.83 & 0.68 & 0.89 \\
        \texttt{LSTM} & 0.94 & 0.75 & 0.60 & 0.82 & 0.96 & 0.81 & 0.68 & 0.87 \\
        \texttt{RoBERTa} & 0.98 & 0.83 & 0.70 & 0.88 & 0.99 & 0.86 & 0.75 & 0.90 \\
    \bottomrule
  \end{tabular}}  
\end{table*}

\subsection{Multi-class Sentiment Analysis Benchmark}
We report model performance results under 3 training conditions: \textbf{Binary Classification}, where we label reviews with 1 star and 2 star ratings as negative, reviews with 4 star and 5 star as positive, and 3-star reviews are dropped; \textbf{Ternary Classification}, where we add another neutral class for reviews with 3 star ratings; and \textbf{5-way Classification}, where each star rating by itself is considered as a class. We leave out reviews in the train set in the `no majority' category. (Dev and Test do not contain any such examples.) Table~\ref{tab:seq-cls-perf} shows the performance results for our models under different conditions. Our results suggest that \texttt{RoBERTa} has the edge over others across all evaluated tasks.  

\subsection{Aspect-based Sentiment Analysis Benchmark}\label{app:absa-models}

Our dataset can be naturally used as an aspect-based sentiment analysis (ABSA) benchmark. For each sentence, it may contain up to 4 aspects with respect to the reviewing restaurant. As ABSA benchmarks are usually small and sparse with missing labels, our dataset provides validated aspect-based labels, and is one of the largest human validated ABSA benchmark.

To evaluate model performance, we adapt standard finetuning approach for ABSA benchmarks as proposed by \cite{sun2019utilizing}. Instead of single sentence classification, we add another auxiliary sentence representing the aspect. For instance, to predict the label for the `food' aspect for ``the food here is good but not the service'', we append a single aspect token with a separator, and construct our input sentence as ``the food here is good but not the service [SEP] food''. 
Table~\ref{tab:seq-cls-perf} shows the performance results for our models under different conditions.

\begin{table*}[tp]
\centering
 \caption{Model performance on top-k author identity prediction with number of train and dev examples.}
  \label{tab:author-id-pred}
  
  \vspace{3pt}
  
\resizebox{0.7\linewidth}{!}{%
  \centering
  \setlength{\tabcolsep}{12pt}
  \begin{tabular}[c]{l c c c c}
    \toprule
        \textbf{Model} & Accuracy & Macro-F1 & \# train & \# dev  \\
    \midrule
        Random (k=5) & 0.16 & 0.15 & 1105 & 227 \\
        Random (k=10) & 0.10 & 0.10 & 2072 & 519 \\
        Random (k=15) & 0.07 & 0.07 & 2963 & 741 \\
    \midrule
        \texttt{RoBERTa} (k=5) & 0.27 & 0.16 & 1105 & 227 \\
        \texttt{RoBERTa} (k=10) & 0.14 & 0.05 & 2072 & 519 \\
        \texttt{RoBERTa} (k=15) & 0.11 & 0.04 & 2963 & 741 \\
    \bottomrule
  \end{tabular}} 
\end{table*}

\subsection{Author Identity Prediction}
One potential artifact of our benchmark is edited sentence may expose author identity, which may result in artifact in interpreting model performance. To quantify this potential artifact, we train models to predict author identities based on the sentences. We create author identity prediction dataset by aggregating our dataset by anonymized worker ids. We then split the dataset into train/dev with a 4-to-1 ratio. For model training, we finetune \texttt{RoBERTa} for 5 epochs with a batch size of 32, a learning rate of $2e^{-5}$, and a maximum sequence length of 128. Note that we only consider top-k annotators ranked by their contributions (i.e., number of examples in our dataset). Table~\ref{tab:author-id-pred} shows the performance results of our finetuned models with a random classifier. Our results suggest that potential artifacts may exist but only for a limited extend.

\section{Additional Results}\label{app:extended-results}

In this section, we report additional results for \texttt{bert-base-uncased}, \texttt{roberta-base}, \texttt{gpt-2}, and an \texttt{LSTM}, fine-tuned on binary, ternary and 5-way versions of the sentiment task. These models are described in Appendix~\ref{app:cebab-models}. Table \ref{tab:cebab-baselines} summarizes all the results.

We refer to the results section in the main text for an explanation of the different metrics considered. Which metric is best depends on the final use-case and whether it is more important to estimate the direction or the magnitude of the effect.

\begin{figure}[ht]
  \centering
  \includegraphics[width=0.8\textwidth]{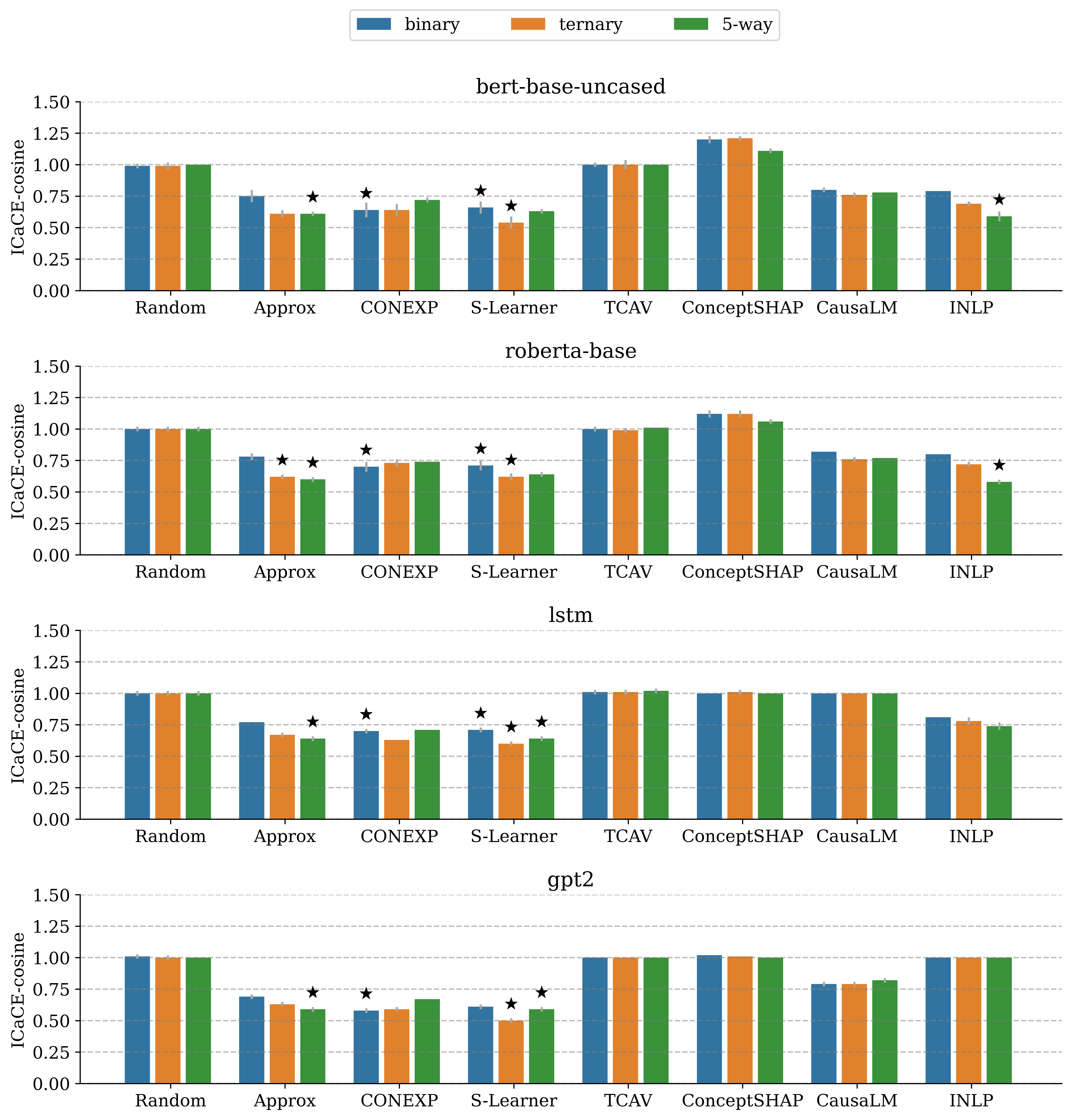}
  \caption{ICaCE-Error for all experiments using the \textit{cosine} distance metric. \textbf{Lower is better}. Results averaged over 5 distinct seeds. Error bars (in gray) display the standard deviation. Stars denote the best results for a given classification setting.}
  \label{fig:icace-cosine}
\end{figure}

\paragraph{ICaCE-cosine}
Figure~\ref{fig:icace-cosine} shows the results for the ICaCE-Error with the \textit{cosine} distance metric. The explanation methods that take the direction of the intervention into account (Approx, CONEXP, S-Learner) are the clear winners across all different models considered. 
S-Learner marginally wins across the most settings, but the conceptually simple Approx baseline is a close second. The strong performance of this simple baseline across the board suggests that most methods perform subpar, and that there is potential value in developing better concept-based model explanation methods.

Both TCAV and ConceptSHAP struggle to achieve better-than-random performance across all settings. Further analysis is needed to exactly understand why these methods are struggling.

Some additional trends emerge that require more analysis to fully understand. For example, Approx generally increases in performance when evaluated on more fine-grained classification settings, while CONEXP is typically worse here.

\begin{figure}[ht]
  \centering
  \includegraphics[width=0.8\textwidth]{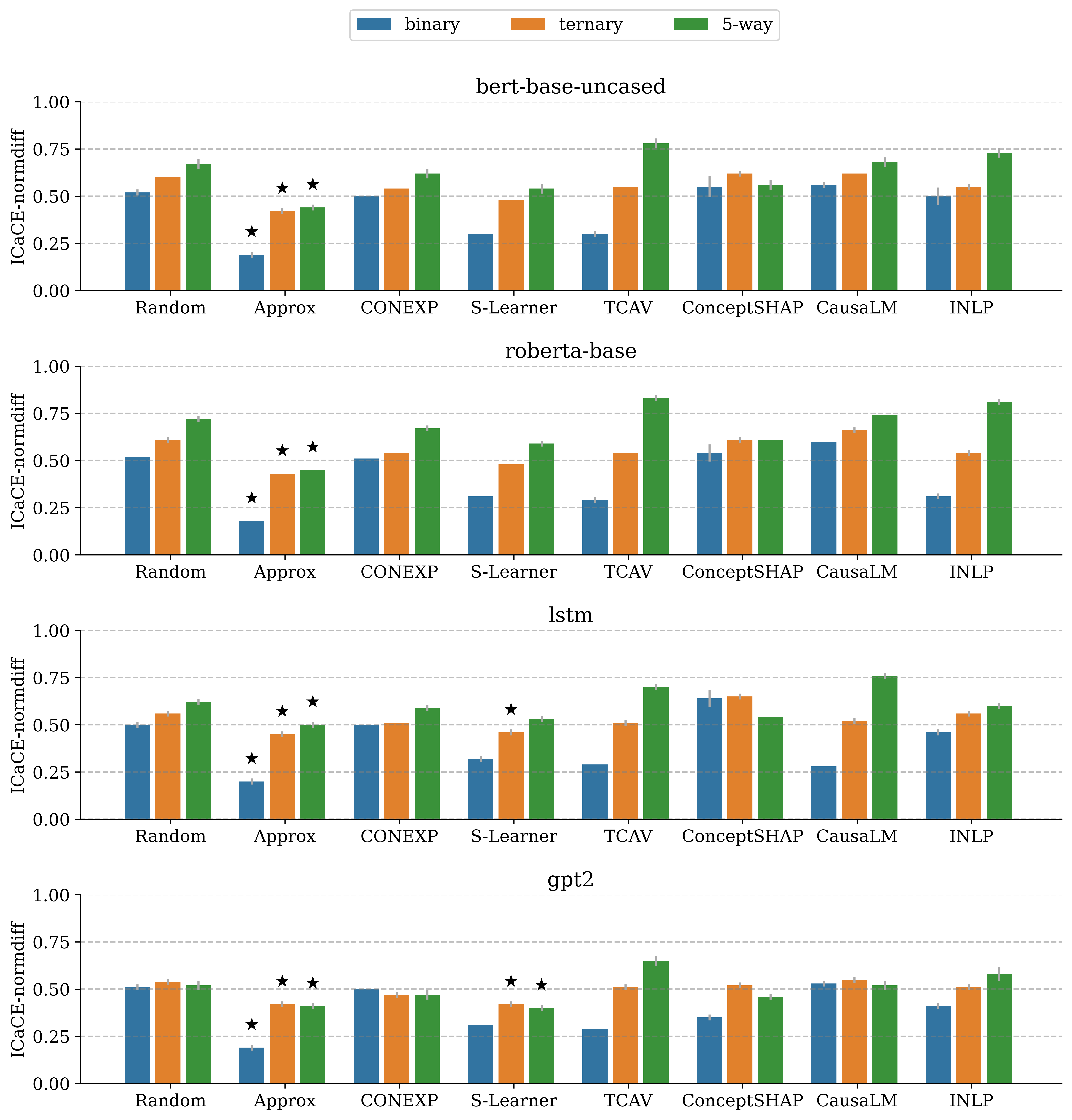}
  \caption{ICaCE-Error for all experiments using the \textit{normdiff} distance metric. \textbf{Lower is better}. Results averaged over 5 distinct seeds. Error bars (in gray) display the standard deviation. Stars denote the best results for a given classification setting.}
  \label{fig:icace-normdiff}
\end{figure}

\paragraph{ICaCE-normdiff}
Figure~\ref{fig:icace-normdiff} shows the results for the ICaCE-Error with the \textit{normdiff} distance metric. In general, it is more difficult for explanation methods to estimate the magnitude of the intervention effect when the task increases in complexity. For a given explanation method and model, best results are often achieved for the binary classification problem.

The conceptually simple Approx baseline wins across the board. S-Learner is only able to match its performance a few times. While previous results already showed that most of the methods fall behind the Approx baseline, the results are particularly striking for this metric.

While S-learner and CONEXP were somewhat comparable on the \textit{cosine} metric, their differences become clear on the \texttt{normdiff} metric: S-Learner is better at estimating the magnitude of the intervention.

An interesting trend can be observed for TCAV, which has good performance on the binary task but becomes worse than random when evaluated on the ternary and 5-way settings. ConceptSHAP is the only method that consistently breaks the upward trend when going from ternary to the 5-way setting. More analysis is needed to understand both these phenomena.

\begin{figure}[ht]
  \centering
  \includegraphics[width=0.8\textwidth]{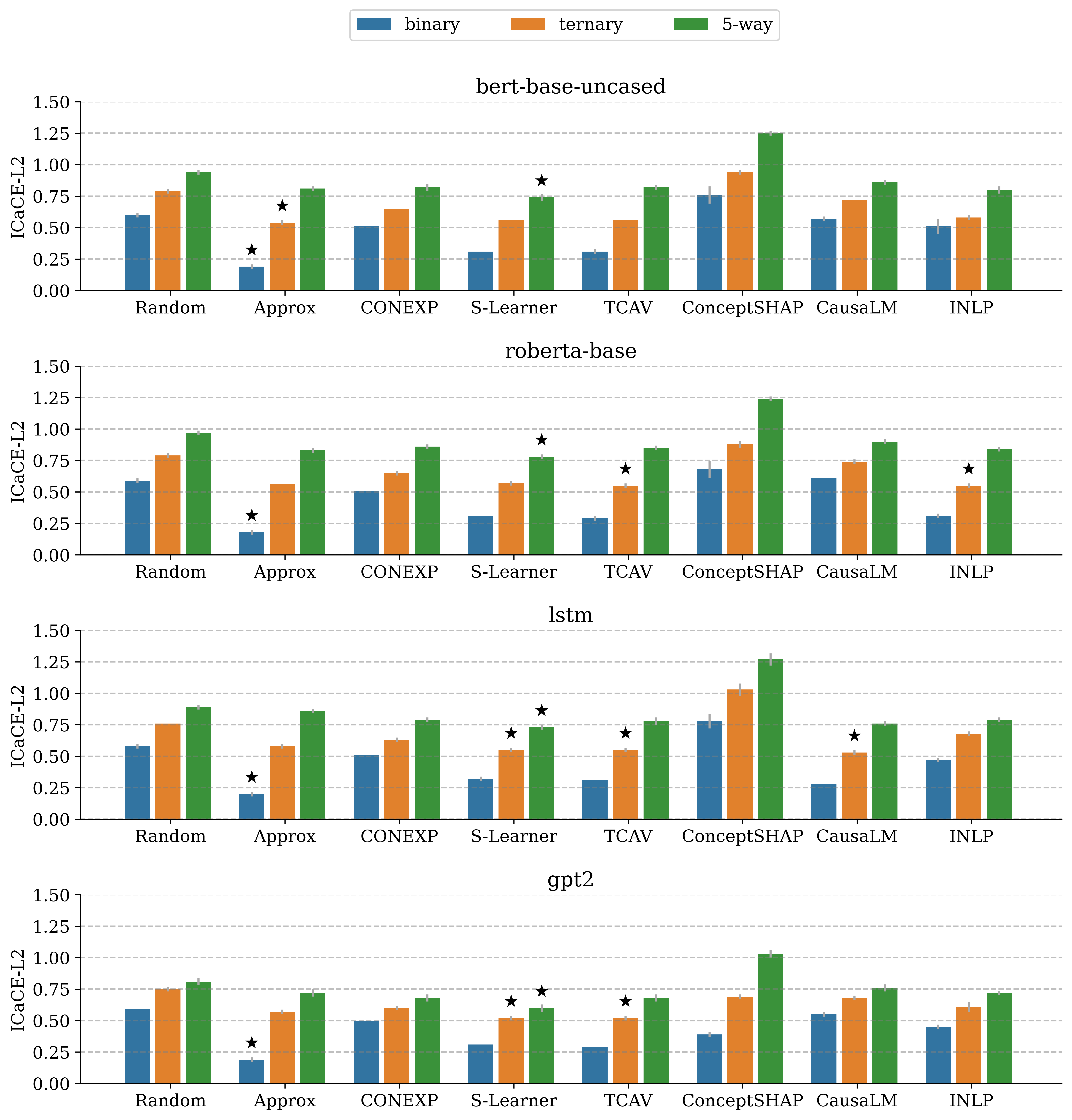}
  \caption{ICaCE-Error for all experiments using the \textit{L2} distance metric. \textbf{Lower is better}. Results averaged over 5 distinct seeds. Error bars (in gray) display the standard deviation. Stars denote the best results for a given classification setting.}
  \label{fig:icace-l2}
\end{figure}

\paragraph{ICaCE-L2}
Figure~\ref{fig:icace-l2} shows the results for the ICaCE-Error with the \textit{L2} distance metric. Because this metric takes both the scale and direction of the effect into account, it is slightly harder to interpret. In general, the performance drops when evaluated on more fine-grained classification settings.

Again, the Approx baseline is a strong contestant, but on this metric the results are more varied. S-Learner is consistently the best at producing the closest explanation in Euclidian distance to the real effect for the 5-way setting.

\begin{table*}[tp]
\caption{ICaCE scores on the test set for the binary, ternary and 5-way classification settings. \textbf{Lower is better}. Results averaged over 5 distinct seeds; standard deviations in parentheses.}
\label{tab:cebab-baselines}
  
\vspace{3pt}
    
\centering

\begin{subtable}{1.0\textwidth}
    \caption{ICaCE scores for 5-way sentiment classification setting.}
    \resizebox{1.0\textwidth}{!}{%
\begin{tabular}{lrcccccccc}
\toprule
\multirow{1}{*}{Model} & \multirow{1}{*}{Metric} & \texttt{Random} & \texttt{Approx} & \texttt{CONEXP} & \texttt{S-Learner} & \texttt{TCAV} & \texttt{ConceptSHAP} & \texttt{CausaLM} & \texttt{INLP} \\
\midrule

\multirow3{*}{\texttt{BERT}}
& $\texttt{L2}_{\text{ICaCE}}$ & 0.94 (.01) & 0.81 (.01) & 0.82 (.02) & 0.74 (.02) & 0.82 (.01) & 1.25 (.01) & 0.86 (.01) & 0.80 (.02) \\
& $\texttt{COS}_{\text{ICaCE}}$ & 1.00 (.00) & 0.61 (.01) & 0.72 (.01) & 0.63 (.01) & 1.00 (.00) & 1.11 (.01) & 0.78 (.00) & 0.59 (.03) \\
& $\texttt{NormDiff}_{\text{ICaCE}}$ & 0.67 (.02) & 0.44 (.01) & 0.62 (.02) & 0.54 (.02) & 0.78 (.02) & 0.56 (.02) & 0.68 (.02) & 0.73 (.02) \\
\midrule
\multirow3{*}{\texttt{RoBERTa}}
& $\texttt{L2}_{\text{ICaCE}}$ & 0.97 (.01) & 0.83 (.01) & 0.86 (.01) & 0.78 (.01) & 0.85 (.01) & 1.24 (.01) & 0.90 (.01) & 0.84 (.01) \\
& $\texttt{COS}_{\text{ICaCE}}$ & 1.00 (.01) & 0.60 (.01) & 0.74 (.00) & 0.64 (.01) & 1.01 (.00) & 1.06 (.01) & 0.77 (.00) & 0.58 (.01) \\
& $\texttt{NormDiff}_{\text{ICaCE}}$ & 0.72 (.01) & 0.45 (.00) & 0.67 (.01) & 0.59 (.01) & 0.83 (.01) & 0.61 (.00) & 0.74 (.00) & 0.81 (.01) \\
\midrule
\multirow3{*}{\texttt{GPT-2}}
& $\texttt{L2}_{\text{ICaCE}}$ & 0.81 (.02) & 0.72 (.02) & 0.68 (.02) & 0.60 (.02) & 0.68 (.02) & 1.03 (.02) & 0.76 (.02) & 0.72 (.01) \\
& $\texttt{COS}_{\text{ICaCE}}$ & 1.00 (.00) & 0.59 (.01) & 0.67 (.00) & 0.59 (.01) & 1.00 (.00) & 1.00 (.00) & 0.82 (.01) & 1.00 (.00) \\
& $\texttt{NormDiff}_{\text{ICaCE}}$ & 0.52 (.02) & 0.41 (.01) & 0.47 (.02) & 0.40 (.01) & 0.65 (.02) & 0.46 (.01) & 0.52 (.02) & 0.58 (.03) \\
\midrule
\multirow3{*}{\texttt{LSTM}}
& $\texttt{L2}_{\text{ICaCE}}$ & 0.89 (.01) & 0.86 (.01) & 0.79 (.01) & 0.73 (.01) & 0.78 (.02) & 1.27 (.04) & 0.76 (.01) & 0.79 (.01) \\
& $\texttt{COS}_{\text{ICaCE}}$ & 1.00 (.01) & 0.64 (.01) & 0.71 (.00) & 0.64 (.01) & 1.02 (.01) & 1.00 (.00) & 1.00 (.00) & 0.74 (.02) \\
& $\texttt{NormDiff}_{\text{ICaCE}}$ & 0.62 (.01) & 0.50 (.01) & 0.59 (.01) & 0.53 (.01) & 0.70 (.01) & 0.54 (.00) & 0.76 (.01) & 0.60 (.01) \\

\bottomrule
\end{tabular}%
}

\end{subtable}
\vspace{3pt}

\begin{subtable}{1.0\textwidth}
    \caption{ICaCE scores for ternary sentiment classification setting.}
    \resizebox{1.0\textwidth}{!}{%
\begin{tabular}{lrcccccccc}
\toprule
\multirow{1}{*}{Model} & \multirow{1}{*}{Metric} & \texttt{Random} & \texttt{Approx} & \texttt{CONEXP} & \texttt{S-Learner} & \texttt{TCAV} & \texttt{ConceptSHAP} & \texttt{CausaLM} & \texttt{INLP} \\
\midrule

\multirow3{*}{\texttt{BERT}}
& $\texttt{L2}_{\text{ICaCE}}$ & 0.79 (.01) & 0.54 (.01) & 0.65 (.00) & 0.56 (.00) & 0.56 (.00) & 0.94 (.01) & 0.72 (.00) & 0.58 (.01) \\
& $\texttt{COS}_{\text{ICaCE}}$ & 0.99 (.02) & 0.61 (.02) & 0.64 (.04) & 0.54 (.04) & 1.00 (.03) & 1.21 (.01) & 0.76 (.01) & 0.69 (.01) \\
& $\texttt{NormDiff}_{\text{ICaCE}}$ & 0.60 (.00) & 0.42 (.01) & 0.54 (.00) & 0.48 (.00) & 0.55 (.00) & 0.62 (.01) & 0.62 (.00) & 0.55 (.01) \\
\midrule
\multirow3{*}{\texttt{RoBERTa}}
& $\texttt{L2}_{\text{ICaCE}}$ & 0.79 (.01) & 0.56 (.00) & 0.65 (.01) & 0.57 (.01) & 0.55 (.01) & 0.88 (.02) & 0.74 (.01) & 0.55 (.01) \\
& $\texttt{COS}_{\text{ICaCE}}$ & 1.00 (.01) & 0.62 (.01) & 0.73 (.02) & 0.62 (.02) & 0.99 (.01) & 1.12 (.02) & 0.76 (.01) & 0.72 (.01) \\
& $\texttt{NormDiff}_{\text{ICaCE}}$ & 0.61 (.01) & 0.43 (.00) & 0.54 (.00) & 0.48 (.00) & 0.54 (.00) & 0.61 (.01) & 0.66 (.01) & 0.54 (.01) \\
\midrule
\multirow3{*}{\texttt{GPT-2}}
& $\texttt{L2}_{\text{ICaCE}}$ & 0.75 (.01) & 0.57 (.01) & 0.60 (.01) & 0.52 (.01) & 0.52 (.01) & 0.69 (.01) & 0.68 (.01) & 0.61 (.03) \\
& $\texttt{COS}_{\text{ICaCE}}$ & 1.00 (.01) & 0.63 (.01) & 0.59 (.01) & 0.50 (.01) & 1.00 (.00) & 1.01 (.00) & 0.79 (.01) & 1.00 (.00) \\
& $\texttt{NormDiff}_{\text{ICaCE}}$ & 0.54 (.01) & 0.42 (.01) & 0.47 (.01) & 0.42 (.01) & 0.51 (.01) & 0.52 (.01) & 0.55 (.01) & 0.51 (.01) \\
\midrule
\multirow3{*}{\texttt{LSTM}}
& $\texttt{L2}_{\text{ICaCE}}$ & 0.76 (.00) & 0.58 (.01) & 0.63 (.01) & 0.55 (.01) & 0.55 (.01) & 1.03 (.04) & 0.53 (.01) & 0.68 (.01) \\
& $\texttt{COS}_{\text{ICaCE}}$ & 1.00 (.01) & 0.67 (.01) & 0.63 (.00) & 0.60 (.01) & 1.01 (.01) & 1.01 (.01) & 1.00 (.00) & 0.78 (.02) \\
& $\texttt{NormDiff}_{\text{ICaCE}}$ & 0.56 (.01) & 0.45 (.01) & 0.51 (.00) & 0.46 (.01) & 0.51 (.01) & 0.65 (.01) & 0.52 (.01) & 0.56 (.01) \\

\bottomrule
\end{tabular}%
}

\end{subtable}
\vspace{3pt}

\begin{subtable}{1.0\textwidth}
    \caption{ICaCE scores for binary sentiment classification setting.}
    \resizebox{1.0\textwidth}{!}{%
\begin{tabular}{lrcccccccc}
\toprule
\multirow{1}{*}{Model} & \multirow{1}{*}{Metric} & \texttt{Random} & \texttt{Approx} & \texttt{CONEXP} & \texttt{S-Learner} & \texttt{TCAV} & \texttt{ConceptSHAP} & \texttt{CausaLM} & \texttt{INLP} \\
\midrule

\multirow3{*}{\texttt{BERT}}
& $\texttt{L2}_{\text{ICaCE}}$ & 0.60 (.01) & 0.19 (.01) & 0.51 (.00) & 0.31 (.00) & 0.31 (.01) & 0.76 (.06) & 0.57 (.01) & 0.51 (.05) \\
& $\texttt{COS}_{\text{ICaCE}}$ & 0.99 (.01) & 0.75 (.04) & 0.64 (.05) & 0.66 (.04) & 1.00 (.01) & 1.20 (.02) & 0.80 (.01) & 0.79 (.00) \\
& $\texttt{NormDiff}_{\text{ICaCE}}$ & 0.52 (.01) & 0.19 (.01) & 0.50 (.00) & 0.30 (.00) & 0.30 (.01) & 0.55 (.05) & 0.56 (.01) & 0.50 (.04) \\
\midrule
\multirow3{*}{\texttt{RoBERTa}}
& $\texttt{L2}_{\text{ICaCE}}$ & 0.59 (.01) & 0.18 (.01) & 0.51 (.00) & 0.31 (.00) & 0.29 (.01) & 0.68 (.06) & 0.61 (.00) & 0.31 (.01) \\
& $\texttt{COS}_{\text{ICaCE}}$ & 1.00 (.01) & 0.78 (.02) & 0.70 (.03) & 0.71 (.03) & 1.00 (.01) & 1.12 (.02) & 0.82 (.00) & 0.80 (.00) \\
& $\texttt{NormDiff}_{\text{ICaCE}}$ & 0.52 (.00) & 0.18 (.00) & 0.51 (.00) & 0.31 (.00) & 0.29 (.01) & 0.54 (.04) & 0.60 (.00) & 0.31 (.01) \\
\midrule
\multirow3{*}{\texttt{GPT-2}}
& $\texttt{L2}_{\text{ICaCE}}$ & 0.59 (.00) & 0.19 (.01) & 0.50 (.00) & 0.31 (.00) & 0.29 (.00) & 0.39 (.01) & 0.55 (.01) & 0.45 (.01) \\
& $\texttt{COS}_{\text{ICaCE}}$ & 1.01 (.01) & 0.69 (.01) & 0.58 (.01) & 0.61 (.01) & 1.00 (.00) & 1.02 (.00) & 0.79 (.01) & 1.00 (.00) \\
& $\texttt{NormDiff}_{\text{ICaCE}}$ & 0.51 (.01) & 0.19 (.01) & 0.50 (.00) & 0.31 (.00) & 0.29 (.00) & 0.35 (.01) & 0.53 (.01) & 0.41 (.01) \\
\midrule
\multirow3{*}{\texttt{LSTM}}
& $\texttt{L2}_{\text{ICaCE}}$ & 0.58 (.01) & 0.20 (.01) & 0.51 (.00) & 0.32 (.01) & 0.31 (.00) & 0.78 (.05) & 0.28 (.00) & 0.47 (.01) \\
& $\texttt{COS}_{\text{ICaCE}}$ & 1.00 (.01) & 0.77 (.00) & 0.70 (.01) & 0.71 (.01) & 1.01 (.01) & 1.00 (.00) & 1.00 (.00) & 0.81 (.00) \\
& $\texttt{NormDiff}_{\text{ICaCE}}$ & 0.50 (.01) & 0.20 (.01) & 0.50 (.00) & 0.32 (.01) & 0.29 (.00) & 0.64 (.04) & 0.28 (.00) & 0.46 (.01) \\

\bottomrule
\end{tabular}%
}

\end{subtable}

\end{table*}

\section{CausaLM}\label{app:causalm}

\subsection{Our adaptation}

The CausaLM algorithm was originally designed to estimate the average treatment effect of a high-level concept on pre-trained language models. Its output estimator is the textual representation averaged treatment effect (TReATE), which is computed as:
\begin{equation}\label{eq:TReATE}
    \text{TReATE}_{\nn_{\phi}}(C; \mathcal{D}) = \frac{1}{|\mathcal{D}|} \sum_{x \in \mathcal{D}}
    \nn' \big(\phi^{\text{CF}}_C(x)\big) -  \nn \big(\phi(x)\big) ,
\end{equation}
where $\phi^{\text{CF}}_C$ denotes the learned counterfactual representation that information about concept $C$ is not present, $\nn'$ is a classifier trained on this counterfactual representation, and $\mathcal{D}$ is a dataset. 

However, for comparison on the \cebab{} data, we require the estimation of individual causal concept effects (ICaCE). To allow a fair comparison, we swap the TReATE output estimator with TReITE (Equation~\ref{eq:TReATE_hat_O_CF}). The only difference between these estimators is that in TReITE we remove the average across $\mathcal{D}$, and output the estimated effect of individual examples.

\subsection{Implementation details}

For all counterfactual models, we optimize using the Adam optimizer with \texttt{lr=2e-5}, \texttt{epochs=3}, \texttt{batch\_size=48}, and the relative weight of the adversarial task, $\lambda$, is set to $0.1$.

For both the factual models and fine-tuning phase, we optimize using the Adam optimizer with \texttt{lr=1e-3}, \texttt{epochs=50}, and \texttt{batch\_size=256}. The differences in hyperparameter values is due to the different architectures we employ; for the counterfactual models we train the entire language model ($\phi$), and for the factual models and the fine-tuning phase we freeze the embedding weights ($\phi$) and train only the classification head ($\mathcal{N}$).

All CausaLM models were trained using 2 Nvidia GTX 1080 Ti 12GB GPUs.

\section{INLP}

\subsection{Our adaptation}

The INLP algorithm was originally designed to debias word embeddings by iteratively projecting them onto the null-space of some protected attribute (concept). However, INLP may serve as an estimation method similar to CausaLM, with the two following crucial differences. First, its lack of ability to control for potential confounders. Second, it operates on the representation rather than on the actual model weights. Since CausaLM and INLP share common characteristics, their output estimators are computed in the same way. See \S\ref{app:causalm} for extended details.

\subsection{Implementation details}

In order to guard for a \say{protected attribute} (concept), INLP determines whether this concept is present in an embedding or not by learning a linear separator in the embedding space. Following the practice suggested in the original paper, we choose our linear separator to be an SVM learned using SGD with $\alpha = 0.01$, $\varepsilon = 0.001$, and \texttt{max\_iter=1000}. Logistic regression showed similar behavior. We project the representation to the null-space with respect to the concept $10$ times. In fact, and similarly to the original paper, we converge to random accuracy of predicting the concept from the counterfactual representation after 4-5 iterations.

For all concepts, the classification head on top of the language model that trained to predict the overall sentiment labels trains for 5 epochs using the Adam optimizer with \texttt{lr=2e-5}.

\section{TCAV}

\subsection{Our adaptation}

The Testing with Concept Activation Vectors (TCAV) explanation method was originally designed to count the percentage of test inputs from dataset $\mathcal{D}$ that are positively influenced by some high-level concept. It outputs a count over the number of examples that are change towards the direction of concept $C$, and computed as:
\begin{equation}\label{eq:original_tcav}
    \text{TCAV}_{\nn_{\phi}}(k, C; \mathcal{D}) = \frac{\left| \left\{ x\in \mathcal{D} : \nabla \nn_k (\phi(x)) \cdot v_C > 0 \right\} \right|}{\left| \mathcal{D} \right|},
\end{equation}
where $k$ is some class index and $v_C$ is a linear direction in the activation space, given by the coefficients of a linear separator trained to distinguish between examples that include or exclude the concept $C$.

While TCAV's output is a count over examples, we use the raw sensitivity (directional derivative). This approach is supported by the authors of the original paper: \say{one could also use a different metric that considers the magnitude of the conceptual sensitivities} \cite{kim_interpretability_2018}.
Also, since TCAV operates on the gradients of a model's logits but the ICaCEs are the difference of two probability vectors, we normalize its outputs by taking \texttt {Tanh}. 

\subsection{Implementation details}\label{app:tcav-implementation}

To learn the Concept Activation Vector (CAV, i.e., a linear direction in the activation space of $\phi$), we train a linear separator to distinguish between examples that include the concept (labeled positive or negative) and examples that do not include it (labeled unknown). When learning CAVs, we drop all \cebab\ train examples that are not labeled for aspect (concept) or do not have a majority with respect to the aspect.

Identically to the original paper, our CAV linear separator is an SVM learned using SGD with $\alpha = 0.01$, $\varepsilon = 0.001$ and \texttt{max\_iter} $= 1000$. 

\section{ConceptSHAP}\label{app:concept-shap}

\subsection{Our adaptation}

The original ConceptSHAP algorithm takes a complete set of concepts $C \in \{C_1, ..., C_m\}$ (such that its completeness score in Equation \ref{eq:completeness} is higher than some threshold) and outputs the relative contribution to the test accuracy of each $C_i$. It outputs an estimator given by the following formula
\begin{equation}\label{eq:shap_score}
    \text{Shapley}_{\{C_1, ..., C_m\}}(C) = 
    \sum_{S \subseteq \{ C_1, \ldots, C_m\} \backslash C} {\frac{\left( m - \left| S \right| -1 \right)!\left| S \right|!}{m!}\left[ \eta (S \cup \left\{ C \right\}) - \eta(S) \right]},
\end{equation}
where $\eta$ is a scoring function operating on sets of concepts that output accuracy ratios.

Similarly to the other methods, if $\eta$ outputs accuracy ratios, then the output of ConceptSHAP is not a suitable estimator for ICaCE. Our straightforward adaptation for ConceptSHAP is to make $\eta$ output class probabilities for classes instead of accuracy ratios. 

Our adapted version outputs a vector for each $C \in \{ C_1, \ldots, C_m\}$ and $x$ according to the following equation:
\begin{equation}\label{concept_shap_score}
    \text{ConceptSHAP}_{\nn_{\phi}}(C; x) = 
    \sum_{S \subseteq \{ C_1, \ldots, C_m\} \backslash C} {\frac{\left( m - \left| S \right| -1 \right)!\left| S \right|!}{m!}\left[ \eta (S \cup \left\{ C \right\}) - \eta(S) \right]},
\end{equation}
where $\eta$ is a function defined as $\eta_{\nn_{\phi}}(S) = \sup_g\nn\big(g\big(V_S \,\phi(x)\big)\big)$, and $V_S$ is a matrix with the learned concept directions as its rows $V_S = \left( v_C^T \right)_{C\in S} \in \mathbb{R}^{|S| \times h}$.

\citet{yeh2020completeness} calculate concept directions $v_{C_j}$ automatically by learning a neural network classifier. To allow for a fair comparison between ConceptSHAP and the other evaluated methods, we use the concept activation vectors $v_{C_1}, \ldots ,v_{C_m}$ as the input concepts (similarly to those used in \citet{kim_interpretability_2018}). 

In addition, in the original paper the authors learn the concepts $v_C$ automatically, by using a carefully constructed loss function. To allow a fair comparison, we learn the concept vector by exploiting our labeled aspects (concepts), in a way similar to TCAV. See Section~\ref{app:tcav-implementation} for more details.

\subsection{Completeness Scores of Treatment Concepts}
Given a feature representation $\phi$ and a classification head $\nn$, the completeness score is defined by:
\begin{equation}\label{eq:completeness}
    \text{completeness}_{\nn_\phi}(S; D, Y) = \frac{\sup_g \frac{1}{|\mathcal{D}|} \sum_{(x, y) \in \mathcal{D}, Y} \mathbbm{1}\left[ y= \arg\max_{y'} \nn_{y'}\big(g\big(V_S \,\phi(x)\big)\big)\right] - a_r}{ \frac{1}{|\mathcal{D}|} \sum_{(x, y) \in \mathcal{D}, Y} \mathbbm{1}\left[ y= \arg\max_{y'} \nn_{y'}\big(\phi(x)\big)\right] - a_r },
\end{equation}
where $a_r$ is is the accuracy of a classifier that outputs random predictions, $S \subseteq \{C_1, ..., C_m\}$ and $V_S$ is a matrix with the learned concept directions as its rows $V_S = \left( v_C^T \right)_{C\in S} \in \mathbb{R}^{|S| \times h}$.

For all models, the completeness we get for the set of concepts $S = \{ \text{ambiance}, \text{food}, \text{service}, \text{noise} \}$ is larger than $0.9$.

\subsection{Hyperparameters}

The hyperparameters for CAV are identical to those of TCAV (Section~\ref{app:tcav-implementation}). To calculate $\eta$ and the completeness score, we follow the original paper and set $g$ to be a two-layer perceptron with 500 hidden units, learned using Adam optimizer for 50 epochs, employing \texttt{lr=1e-2} and \texttt{batch\_size=128}.

\end{document}